\documentclass[12pt]{article}
\usepackage[utf8]{inputenc}
\usepackage{graphicx,amsmath,amsfonts,amssymb,fullpage,url,natbib,setspace}
\usepackage{mlmodern}
\usepackage[T1]{fontenc}
\usepackage{xcolor}
\usepackage{hyperref}
\usepackage{bookmark}

\title{\textbf{Predictive representations: \\building blocks of intelligence}}
\author{Wilka Carvalho,$^{1\ast}$ Momchil S. Tomov,$^{2,3\ast}$ William de Cothi,$^{4\ast}$ \\Caswell Barry,$^4$ Samuel J. Gershman$^{1,2,5}$\\
$^1$Kempner Institute for the Study of Intelligence, Harvard University\\
$^2$Department of Psychology and Center for Brain Science, Harvard University\\
$^3$Motional AD LLC\\
$^4$Department of Cell and Developmental Biology, University College London\\
$^5$Center for Brains, Minds, and Machines, MIT \\
$^\ast$Equal contribution}
\date{}

\newcommand{\argmax}{\operatornamewithlimits{argmax}}

\renewcommand{\eqref}[1]{Eq.~\ref{#1}}
\newcommand{\Eqref}[1]{Eq.~\ref{#1}}

\newcommand{\envprobs}{T}
\newcommand{\task}{\mathbf{w}}
\newcommand{\behavior}{\mathbf{z}_{\pi}}
\newcommand{\otherstate}{\tilde{s}}
\newcommand{\rlexp}[1]{\mathbb{E} \left[#1\right]}

\newcommand{\sr}{M}
\renewcommand{\sf}{\boldsymbol{\psi}}

\newcommand{\cumulant}{\boldsymbol{\phi}}
\newcommand{\horizon}{\boldsymbol{\mu}}

\newcommand{\otherstatemarginal}{p}

\definecolor{myred}{HTML}{E06666}
\definecolor{myblue}{HTML}{299DD3}
\definecolor{mygreen}{HTML}{159B27}

\begin{document}

\maketitle

\begin{abstract}
    Adaptive behavior often requires predicting future events. The theory of reinforcement learning prescribes what kinds of predictive representations are useful and how to compute them. This paper integrates these theoretical ideas with work on cognition and neuroscience. We pay special attention to the \emph{successor representation} (SR) and its generalizations, which have been widely applied both as engineering tools and models of brain function. This convergence suggests that particular kinds of predictive representations may function as versatile building blocks of intelligence.
\end{abstract}

\newpage
\tableofcontents

\section{Introduction}

The ability to make predictions has been hailed as a general feature of both biological and artificial intelligence, cutting across disparate perspectives on what constitutes intelligence \citep[e.g.,][]{ciria21,clark13,friston09,ha18,hawkins04,littman01,lotter16}. Despite this general agreement, attempts to formulate the idea more precisely raise many questions: Predict what, and over what timescale? How should predictions be represented? How should they be used, evaluated, and improved? These normative ``should'' questions have corresponding empirical questions about the nature of prediction in biological intelligence. Our goal is to provide systematic answers to these questions. We will develop a small set of principles that have broad explanatory power.

Our perspective is based on an important distinction between predictive \emph{models} and predictive \emph{representations}. A predictive model is a probability distribution over the dynamics of a system's state. A model can be ``run forward'' to generate predictions about the system's future trajectory. This offers a significant degree of flexibility: an agent with a predictive model can, given enough computation time, answer virtually any query about the probabilities of future events. However, the ``given enough computation time'' proviso places a critical constraint on what can be done with a predictive model in practice. An agent that needs to act quickly under stringent computational constraints may not have the luxury of posing arbitrarily complex queries to its predictive model. Predictive representations, on the other hand, cache the answers to certain queries, making them accessible with limited computational cost.\footnote{While we adhere to this definition consistently throughout the paper, we recognize that other usages of the phrase ``predictive representation'' appear in the literature.} The price paid for this efficiency gain is a loss of flexibility: only certain queries can be accurately answered.

Caching is a general solution to ubiquitous flexibility-efficiency trade-offs facing intelligent systems \citep{dasgupta21}. Key to the success of this strategy is caching representations that make task-relevant information directly accessible to computation. We will formalize the notion of task-relevant information, as well as what kinds of computations access and manipulate this information, in the framework of reinforcement learning (RL) theory \citep{sutton18}. In particular, we will show how one family of predictive representation, the \emph{successor representation} (SR) and its generalizations, distills information that is useful for efficient computation across a wide variety of RL tasks. These predictive representations facilitate exploration, transfer, temporal abstraction, unsupervised pre-training, multi-agent coordination, creativity, and episodic control. On the basis of such versatility, we argue that these predictive representations can serve as fundamental building blocks of intelligence.

Converging support for this argument comes from cognitive science and neuroscience. We review a body of data indicating that the brain uses predictive representations for a range of tasks, including decision making, navigation, and memory. We also discuss biologically plausible algorithms for learning and computing with predictive representations. This convergence of biological and artificial intelligence suggests that predictive representations may be a widely used tool for intelligent systems.

Several previous surveys on predictive representations have scratched the surface of these connections \citep{gershman18,momennejad20}. The purpose of this survey is to approach the topic in much greater detail, yielding a comprehensive reference on both technical and scientific aspects. Despite this broad scope, the survey's focus is restricted to predictive representations in the domain of RL; we do not review predictive representations that have been developed for language modeling, vision, and other problems. An important long-term goal will be to fully synthesize the diverse notions of predictive representations across these domains.

\section{Theory}\label{sec:theory}

In this section, we introduce the general problem setup and a classification of solution techniques. We then formalize the SR and discuss how it fits into the classification scheme. Finally, we describe two key extensions of the SR that make it much more powerful: the successor model and successor features. Due to space constraints, we omit some more exotic variants such as the first-occupancy representation~\citep{moskovitz2021first} or the forward-backward representation~\citep{touati2022does}.

\subsection{The reinforcement learning problem}

We consider an agent situated in a \emph{Markov decision process} (MDP) defined by the tuple $M = (\gamma, \mathcal{S}, \mathcal{A}, T, R)$, where $\gamma \in [0,1)$ is a discount factor, $\mathcal{S}$ is a set of states (the state space), $\mathcal{A}$ is a set of actions (the action space), $T(s'|s,a)$ is the probability of transitioning from state $s$ to state $s'$ after taking action $a$, and $R(s)$ is the expected reward in state $s$.\footnote{For notational convenience, we will assume that the state and action spaces are both discrete, but this assumption is not essential for many of the algorithms described here.}
Following~\citet{sutton18}, we consider settings where the MDP and an agent give rise to a \textit{trajectory} of experience
\begin{equation}
  s_0, a_0, r_1, s_1, a_1, \ldots
\end{equation}
where state $s_t$ and action $a_t$ lead to reward $r_{t+1}$ and state $s_{t+1}$. The agent chooses actions probabilistically according to a state-dependent policy $\pi(a|s)$.

We consider settings where the agent prioritizes immediate reward over future reward, as formalized by the concept of \textit{discounted return}. The \emph{value} of a policy is the expected\footnote{To simplify notation, we will sometimes leave implicit the distributions over which the expectation is being taken.} discounted return:
\begin{align}\label{eqn:value}
    V^\pi(s) = \mathbb{E}\left[ \sum_{t=0}^H \gamma^t R(s_{t+1}) \bigg| s_0 = s \right],
\end{align}
One can also define a \textit{state-action} value (i.e., the expected discounted return conditional on action $a$ in state $s$) by:
\begin{align}\label{eqn:q-value}
    Q^\pi(s,a) = \mathbb{E}\left[ \sum_{t=0}^H \gamma^t R(s_{t+1}) \bigg| s_0 = s, a_0 = a \right]
\end{align}
The optimal policy $\pi^\ast$ is then defined by:
\begin{align}
    \pi^\ast(\cdot|s) = \argmax_{\pi(\cdot|s)} V^\pi(s) = \argmax_{\pi(\cdot|s)} \sum_a \pi(a|s) Q^\pi(s,a),
\end{align}
The optimal policy for an MDP is always deterministic (choose the value-maximizing action):
\begin{align}
\textstyle
    &\pi^\ast(a|s) = \mathbb{I}[a = \argmax_{\tilde{a}} Q^\ast(s,\tilde{a})],
\end{align}
where $\mathbb{I}[\cdot]=1$ if its argument is true, 0 otherwise. This assumes that the agent can compute the optimal values $Q^\ast(s,a)=\max_{\pi} Q^\pi(s,a)$ exactly. In most practical settings, values must be approximated. In these cases, stochastic policies are useful (e.g., for exploration), as discussed later.

The Markov property for MDPs refers to the conditional independence of the past and future given the current state and action. This property allows us to write the value function in a recursive form known as the \emph{Bellman equation} \citep{bellman57}:
\begin{align}\label{eq:value-bellman}
    V^\pi(s) &= \sum_a \pi(a|s) \sum_{s'} T(s'|s,a) \left[ R(s') + \gamma V^\pi(s') \right] \nonumber \\
    &= \mathbb{E}[R(s') + \gamma V^\pi(s')].
\end{align}
Similarly, the state-action value function obeys a Bellman equation:
\begin{align}\label{eq:q-bellman}
    Q^\pi(s,a) = \mathbb{E}[R(s') + \gamma Q^\pi(s',a')].
\end{align}
These Bellman equations lie at the heart of many efficient RL algorithms, as we discuss next.

\subsection{Classical solution methods}\label{sec:algorithmic}

\begin{figure}
\centering
\includegraphics[width=\textwidth]{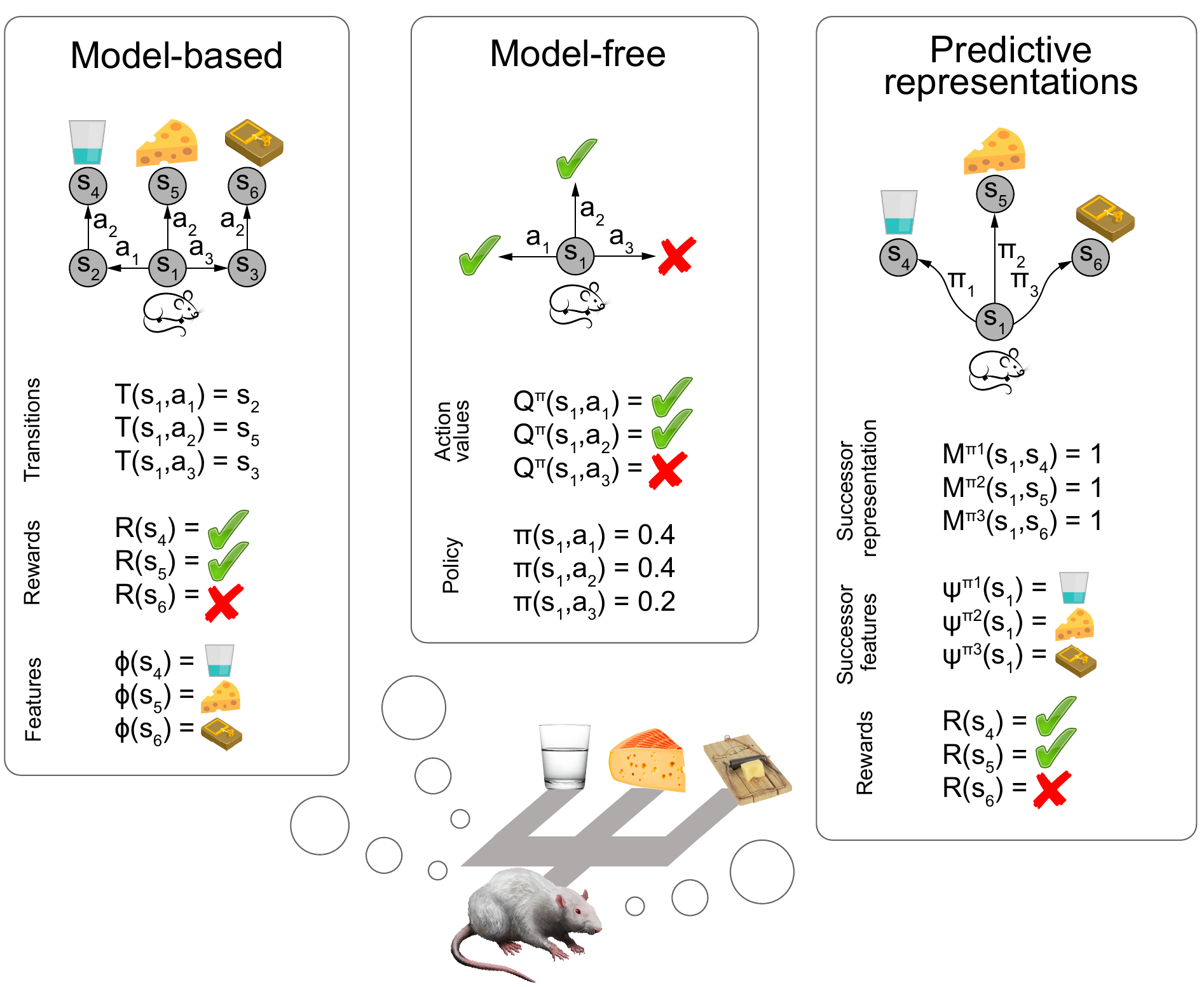}
\caption{
  \textbf{Algorithmic solutions to the RL problem}. An agent solving a three-armed maze (bottom) can adopt different classes of strategies (top). Model-based strategies (left) learn an internal model of the environment, including the transition function ($T$), the reward function ($R$), and (optionally) the features ($\cumulant$). At decision time, the agent can run forward simulations to predict the outcomes of different actions. Model-free strategies (middle) learn action values ($Q$) and/or a policy ($\pi$). At decision time, the agent can consult the cached action values and/or policy in the current state. Strategies relying on predictive representations (right) learn the successor representation (SR) matrix ($\mathbf{M}$) mapping states to future states and/or the successor features ($\sf$) mapping states to future features, as well as the reward function ($R$). At decision time, the agent can consult the cached predictions and cross-reference them with its task (specified by the reward function) to choose an action.
  }
  \label{fig:rl}
\end{figure}

We say that an algorithm solves an MDP if it outputs an optimal policy (or an approximation thereof). Broadly speaking, algorithms can be divided into two classes:
\begin{itemize}
    \item \textbf{Model-based algorithms} use an internal model of the MDP to compute the optimal policy (Figure~\ref{fig:rl}, left).
    \item \textbf{Model-free algorithms} compute the optimal policy by interacting with the MDP (Figure~\ref{fig:rl}, middle).
\end{itemize}
These definitions allow us to be precise about what we mean by \emph{predictive model} and \emph{predictive representation}. A predictive model corresponds to $\hat{T}$, the internal model of the transition distribution, and $\hat{R}$, an internal model of the reward function. An agent equipped with $\hat{T}$ can simulate state trajectories and answer arbitrary queries about the future. Of principal relevance to solving MDPs is policy evaluation---an answer to the query, ``How much reward do I expect to earn in the future under my current policy?'' A simple (but inefficient) way to do this, known as \emph{Monte Carlo policy evaluation}, is by running many simulations from each state (\emph{roll-outs}) and then averaging the discounted return. The basic problem with this approach stems from the \emph{curse of dimensionality} \citep{bellman57}: the trajectory space is very large, requiring a number of roll-outs that is exponential in the trajectory length.

A better model-based approach exploits the Bellman equation. For example, the \emph{value iteration} algorithm starts with an initial estimate of the value function, $\hat{V}^\pi$, then simultaneously improves this estimate and the policy by applying the following update (known as a \emph{Bellman backup}) to each state:
\begin{align}\label{eq:value-fn}
    \hat{V}^\pi(s) \leftarrow \max_a \sum_{s'} \hat{T}(s'|s,a) \left[\hat{R}(s') + \gamma \hat{V}^\pi(s') \right].
\end{align}
This is a form of dynamic programming, guaranteed to converge to the optimal solution, $V^\ast(s) = \max_\pi V^\pi(s)$, when the agent's internal model is accurate ($\hat{T} =T, \hat{R}=R$). After convergence, the optimal policy for state $s$ is given by:
\begin{align}
    &\pi^\ast(a|s) = \mathbb{I}[a = a^\ast(s)], \\
    &a^\ast(s) = \argmax_a  Q^\ast(s,a) \approx \argmax_a  \hat{Q}^\ast(s,a).
\end{align}
The approximation becomes an equality when the agent's internal model is accurate.

Value iteration is powerful, but still too cumbersome for large state spaces, since each iteration requires $\mathcal{O}(|\mathcal{A}||\mathcal{S}|^2)$ steps. The basic problem is that algorithms like value iteration attempt to compute the optimal policy for \emph{every} state, but in an online setting an agent only needs to worry about what action to take in its \emph{current} state. This problem is addressed by tree search algorithms, which rely on roll-outs (as in Monte Carlo policy evaluation), but only from the current state. When combined with heuristics for determining which roll-outs to perform (e.g., learned value functions; see below), this approach can be highly effective \citep{silver16}.

Despite their effectiveness for certain problems (e.g., games like Go and chess), model-based algorithms have had only limited success in a wider range of problems (e.g., video games) due to the difficulty of learning a good model and planning in complex (possibly infinite/continuous) state spaces.\footnote{Recent work on applying model-based approaches to video games has seen some success \citep{tsividis21}, but progress towards scalable and generally applicable versions of such algorithms is still in its infancy.} For this reason, much of the work in modern RL has focused on model-free algorithms.

A model-free agent by definition has no access to $\hat{T}$ (and sometimes no access to $\hat{R}$), but nonetheless can still answer certain queries about the future if it has cached a predictive representation. For example, an agent could cache an estimate of the state-action value function, $\hat{Q}^\pi(s,a) \approx Q^\pi(s,a)$. This predictive representation does not afford the same flexibility as a model of the MDP, but it has the advantage of caching, in a computationally convenient form, exactly the information about the future that an agent needs to act optimally.

Importantly, $\hat{Q}^\pi(s,a)$ can be learned purely from interacting with the environment, without access to a model. For example, \emph{temporal difference} (TD) learning methods use stochastic approximation of the Bellman backup. Q-learning is the canonical algorithm of this kind:
\begin{align}
    &\hat{Q}^\pi(s,a) \leftarrow \hat{Q}^\pi(s,a) + \eta \delta   \label{eq:Qlearn} \\
    &\delta = R(s') + \gamma \max_{a'} \hat{Q}^\pi(s',a') - \hat{Q}^\pi(s,a),
\end{align}
where $\eta \in [0,1]$ is a learning rate and $s'$ is sampled from $T(s'|s,a)$. When the estimate is exact, $\mathbb{E}[\delta] = 0$ and $\hat{Q}^\pi = Q^\pi$. Moreover, these updates converge to $Q^\ast(s,a)$ with probability 1 provided that the learning rates satisfy the standard Robbins-Monro conditions for stochastic approximation \citep{watkins92}.

\begin{figure}
\centering
\includegraphics[width=\textwidth]{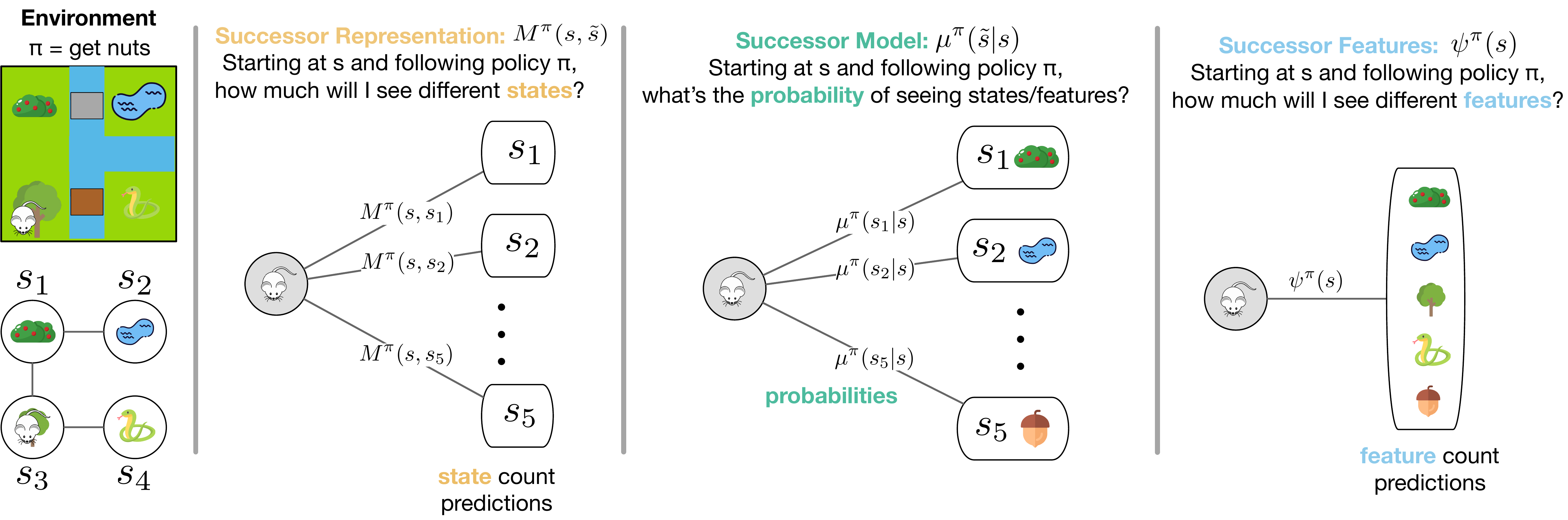}
\caption{Three kinds of predictive representations: the successor representation (\S\ref{sec:sr}), the successor model (\S\ref{sec:successor-models}), and successor features (\S\ref{sec:sf}). Their computations are summarized in Table \ref{table:sr-summary}.
Each of these predictive representations describes a state by a prediction of what will happen when a policy $\pi$ is followed. 
With the \textbf{successor representation}, one gets a description of how much \textit{all} states will be visited in the near future when beginning at state $s$. One limitation of this is that it does not scale well to large state-spaces, since it is impractical to maintain predictions about all states.
\textbf{Successor models} circumvent this challenge by framing learning as a density estimation problem. This enables scaling to high-dimensional state- and action-spaces (including continuous spaces) with amortized learning procedures (\S\ref{sec:learning-horizon}).
\textbf{Successor features} are another method for circumventing the challenge of representing large state-spaces. Here, we do so by describing states with a shared set of state-features and make predictions about how much features will be experienced.
Both successor models and successor features have pros and cons. 
Successor models open up new possibilities like supporting temporally abstract sampling of future states under a policy. Additionally, algorithms for learning successor models typically subsume learning of state-features, whereas successor features typically need a separate mechanism for learning state features. On the other hand, successor features are easier to formulate and more readily enable stitching together policies concurrently (\S\ref{sec:sf-gpi}) and sequentially (\S\ref{sec:ok}) in time---though there is progress on doing this with successor models (see \S\ref{sec:transfer-advances}).
}\label{fig:sr-sf-sm}
\end{figure}

We have briefly discussed the dichotomy of model-based versus model-free algorithms for learning an optimal policy. Model-based algorithms are more flexible---capable of generating predictions about future trajectories---while model-free algorithms are more computationally efficient---capable of rapidly computing the approximate value of an action. The flexibility of model-based algorithms is important for \emph{transfer}: when the environment changes locally (e.g., a route is blocked, or the value of a state is altered), an agent's model will typically also change locally, allowing it to transfer much of its previously learned knowledge without extensive new learning. In contrast, a cached value function approximation (due to its long-term dependencies) will change non-locally, necessitating more extensive learning to update all the affected cached values.

One of the questions we aim to address is how to get some aspects of model-based flexibility without learning and computing with a predictive model. This leads us to another class of predictive representations: the SR. In this section, we describe the SR (\S\ref{sec:sr}), its probabilistic variant (the successor model; \S\ref{sec:successor-models}), and an important generalization (successor features; \S\ref{sec:sf}). A visual overview of these predictive representations is shown in Figure \ref{fig:sr-sf-sm}. Applications of these concepts will be covered in \S\ref{sec:applications}.

\subsection{The successor representation}\label{sec:sr}

The SR, denoted $\sr^{\pi}$, was introduced to address the transfer problem described in the previous section \citep{dayan93,gershman18}. In particular, the SR is well-suited for solving sets of tasks that share the same transition structure but vary in their reward structure; we will delve more into this problem setting later when we discuss applications.

\begin{figure}
\centering
\includegraphics[width=\textwidth]{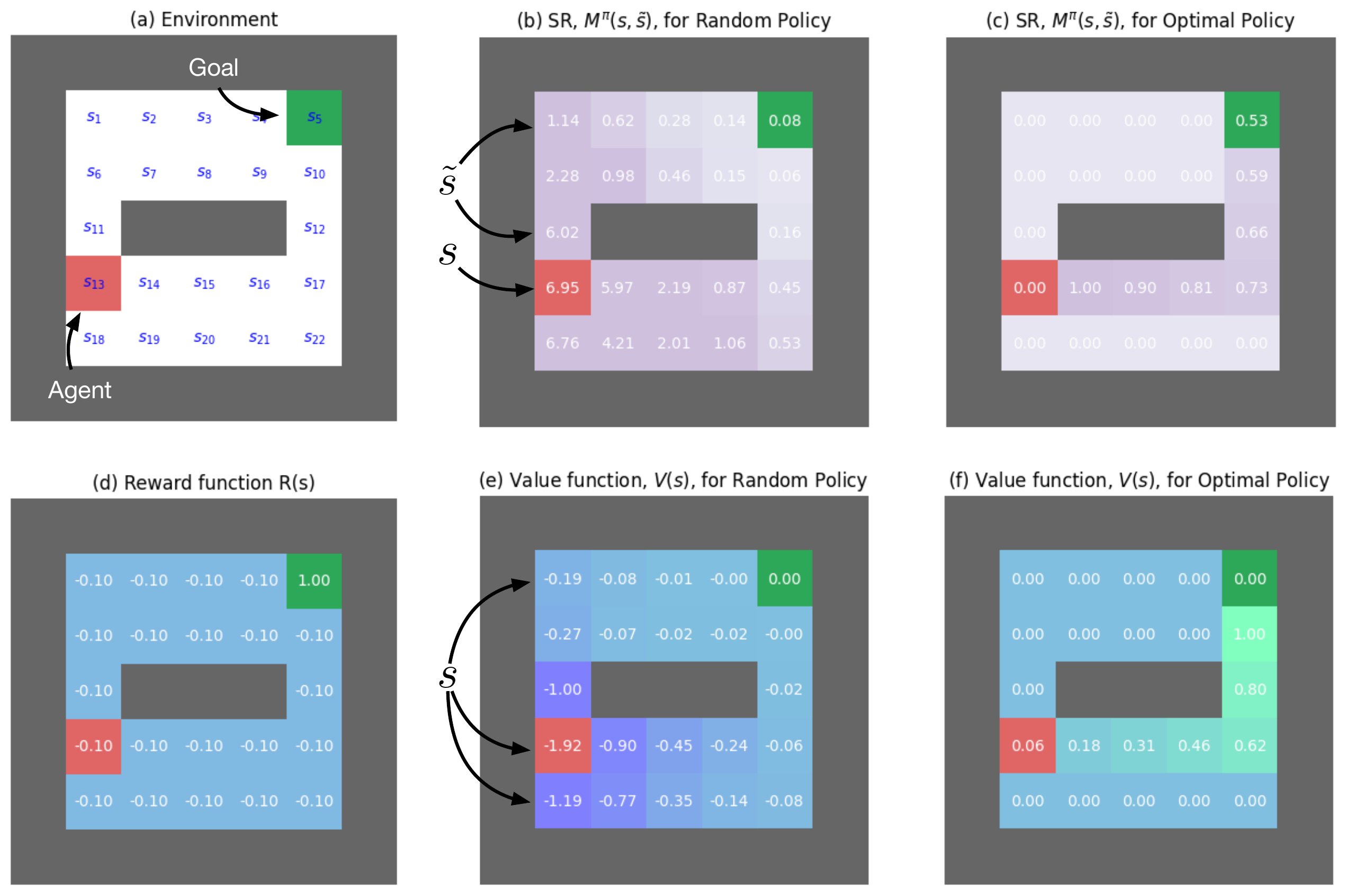}
\caption{
  \textbf{The successor representation (SR).} (a) A schematic of an environment where the agent is a red box at state $s_{13}$ and the goal is a green box at state $s_5$. In general, an SR $\sr^{\pi}(s, \tilde{s})$ (\eqref{eq:sr}) describes the discounted state occupancy for state $\otherstate$ when beginning at state $s$ and following policy $\pi$. In panels (b-c), we showcase $\sr^{\pi}(s_{13}, \tilde{s})$ for a random policy and an optimal policy. (b) The SR under a random policy measures high state occupancy near the agent's current state (e.g., $\sr^{\pi}(s_{13}, s_{14}) = 5.97$) and low state occupancy at points further away from the agent (e.g., $\sr^{\pi}(s_{13}, s_{12}) = 0.16$). (c) The SR under the optimal policy has highest state occupancy along the shortest path to the goal (e.g., here $\sr^{\pi}(s_{13}, s_{12}) = .66$), fading as we get further from the current state. In contrast to a random policy, states not along that path have $0$ occupancy (e.g., $\sr^{\pi}(s_{13}, s_{19}) = 0.0$). Once we know a reward function, we can efficiently evaluate both policies (\eqref{eq:sr-value}). (d) An example reward function that has a cost of $-0.1$ for each state except the goal state where reward is $1$. The SR allows us to efficiently compute (e) the value function under a random policy and (f) the value function under the optimal policy.
  }
  \label{fig:sr-env}
\end{figure}

Like the value function, the SR is a discounted sum over a quantity of interest known as its \emph{cumulant}. The value function has reward as its cumulant, whereas the SR has \emph{state occupancy} as its cumulant:
\begin{equation}\label{eq:sr}
  \sr^\pi\left(s, \otherstate\right) = \mathbb{E}\left[ \sum_{t=0}^H \gamma^t \mathbb{I}\left[s_{t+1}=\otherstate\right] \bigg| s_0=s \right]
\end{equation}
Here $\tilde{s}$ denotes a future state, and $M^\pi(s,\tilde{s}) \in \mathbb{R}$ is the expected discounted future occupancy of $\tilde{s}$ starting in state $s$ under policy $\pi$. An illustration of the SR and comparison with the value function is shown in Figure \ref{fig:sr-env}.

If we define the marginal transition matrix $\mathbf{T}^\pi \in \mathbb{R}^{|S| \times |S|}$ by:
\begin{align}
    T^\pi(s,s') = \sum_a \pi(a|s) T(s'|s,a),
\end{align}
then the SR can be derived analytically from the transition function when $H = \infty$ as:
\begin{align}\label{eq:SRmatrix}
    \mathbf{M}^\pi = \sum_{t=0}^\infty \gamma^t [\mathbf{T}^\pi]^{t+1} = \mathbf{T}^\pi(\mathbf{I} - \gamma \mathbf{T}^\pi)^{-1}
\end{align}
where $\mathbf{M}^\pi \in \mathbb{R}^{|S| \times |S|}$.
\Eqref{eq:SRmatrix} makes explicit the sense in which the SR (a predictive representation) is a compilation of the transition matrix (a predictive model). The SR discards information about individual transitions, replacing them with their cumulants, analogous to how the value function replaces individual reward sequences with their cumulants.

Like the value function, the SR obeys a Bellman equation:
\begin{align}
    M^\pi(s,\otherstate) &= \sum_a \pi(a|s) \sum_{s'} T(s'|s,a) \big[ \mathbb{I}[s'=\otherstate] + \gamma M^\pi(s',\otherstate) \big] \nonumber \\
    &= \mathbb{E}\bigg[\mathbb{I}[s'=\otherstate] + \gamma \sr^\pi(s',\otherstate)\bigg].
\end{align}
This means that it is possible to learn the SR using TD updates similar to the ones applied to value learning:
\begin{align}\label{eq:SRTDupdate}
    \hat{M}^\pi(s,\otherstate) \leftarrow \hat{\sr}^\pi(s,\otherstate) + \eta \delta_M (\otherstate),
\end{align}
where
\begin{align}\label{eq:SRTDerror}
    \delta_M(\otherstate) = \mathbb{I}[s'=\otherstate] + \gamma \hat{M}^\pi(s',\otherstate) - \hat{M}^\pi(s,\otherstate)
\end{align}
is the TD error. Notice that, unlike in TD learning for value, the error is now vector-valued (one error for each state). Once the SR is learned, the value function for a particular reward function under the policy $\pi$ can be efficiently computed as a linear function of the SR:
\begin{equation}\label{eq:sr-value}
  V^\pi(s)=\sum_{\otherstate} \sr^\pi\left(s, \otherstate\right) R\left(\otherstate\right).
\end{equation}
Intuitively,~\Eqref{eq:sr-value} expresses a decomposition of future reward into immediate reward in each state and the frequency with which those states are visited in the near future.

Just as one can condition a value function on actions to obtain a state-action value function (\eqref{eqn:q-value}), we can condition the SR on actions as well\footnote{Note that we overload $\sr^{\pi}$ to also accept actions to reduce the amount of new notation. In general, $\sr^\pi\left(s,  \otherstate\right) = \sum_a \pi(a|s) \sr^\pi\left(s, a, \otherstate\right)$.}:
\begin{align}\label{eq:sr-action}
  \sr^\pi\left(s, a, \otherstate\right) &= \mathbb{E}\left[ \sum_{t=0}^H \gamma^t \mathbb{I}\left[s_{t+1}=\otherstate\right] \bigg| s_0=s, A_0=a \right] \\
  &= \mathbb{E}\bigg[\mathbb{I}[s'=\otherstate] + \gamma \sr^\pi(s', a', \otherstate)\bigg| s_0=s, A_0=a \bigg].
\end{align}
Given an action-conditioned SR, the action value function for a particular reward function can be computed as a linear function of the SR with
\begin{equation}
  Q^{\pi}(s,a) = \sum_{\otherstate} \sr^\pi\left(s, a, \otherstate\right) R(\otherstate).
\end{equation}
Having established some mathematical properties of the SR, we can now explain why it is useful. First, the SR, unlike model-based algorithms, obviates the need to simulate roll-outs or iterate over dynamic programming updates, because it has already compiled transition information into a convenient form: state values can be computed by simply taking the inner product between the SR and the immediate reward vector. Thus, SR-based value computation enjoys efficiency comparable to model-free algorithms.

Second, the SR can, like model-based algorithms, adapt quickly to certain kinds of environmental changes. In particular, local changes to an environment's reward structure induce local changes in the reward function, which immediately propagate to the value estimates when combined with the SR.\footnote{The attentive reader will note that, at least initially, these are not exactly the correct value estimates, because the SR is policy-dependent, and the policy itself requires updating, which may not happen instantaneously (depending on how the agent is optimizing its policy). Nonetheless, these value estimates will typically be an improvement---a good first guess. As we will see, human learning exhibits similar behavior.} Thus, SR-based value computation enjoys flexibility comparable to model-based algorithms, at least for changes to the reward structure. Changes to the transition structure, on the other hand, require more substantial non-local changes to the SR due to the fact that an internal model of the detailed transition structure is not available.

Our discussion has already indicated several limitations of the SR. First, the policy-dependence of its predictions limits its generalization ability. Second, the SR assumes a finite, discrete state space. Third, it does not generalize to new environment dynamics. When the transition structure changes, the~\Eqref{eq:SRmatrix} no longer holds.
We will discuss how the first and second challenges can be addressed in \S\ref{sec:learning} and some attempts to address the third challenge in \S\ref{sec:transfer-advances}.

\subsection{Successor models: a probabilistic perspective on the SR}\label{sec:successor-models}
\begin{figure}
\centering
\includegraphics[width=\textwidth]{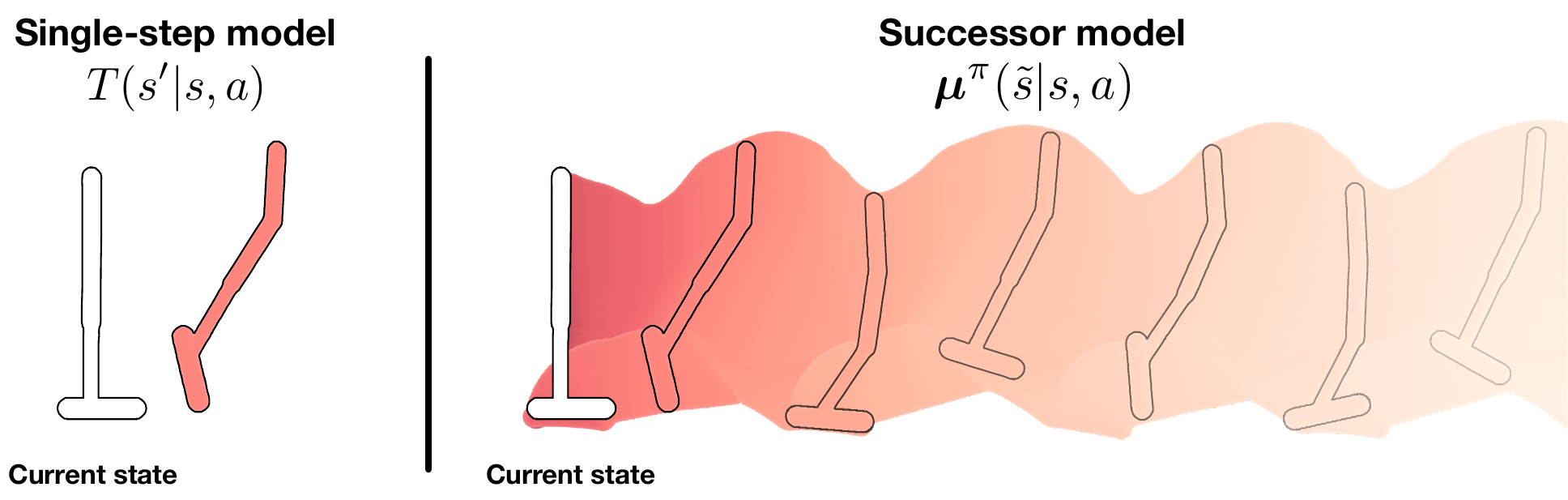}
\caption{
  \textbf{The Successor Model (SM)}. A cartoon schematic of a robot leg that can hop forward. 
  \textbf{Left}: a single-step model can only compute likelihoods for states at the next time-step.
  \textbf{Right}: multi-step successor models can compute likelihoods for states over some horizon into the future.
  One key difference between the SM and the SR is that the SM defines a valid probability distribution. This means that we can leverage density estimation techniques for learning it over continuous state- and action-spaces. Additionally, as this figure suggests, we can use it to sample potentially distal states (see \S\ref{sec:mbrl}).
  Adapted with permission from~\citep{janner2020gamma}.
  }
  \label{fig:successor-models}
\end{figure}

As we've discussed, the SR buys efficiency by caching transition structure, while maintaining some model-based flexibility. One thing that is lost, however, is the ability to simulate trajectories through the state space. In this section, we introduce a generalization of the SR---the \emph{successor model} \citep[SM;][]{janner2020gamma,eysenbach2020c}---that defines an explicit distribution over temporally abstract trajectories (Figure \ref{fig:successor-models}). Temporal abstraction here means a conditional distribution over future states within some time horizon, rather than only the next time-step captured by the transition model.

The SM uses a $k$-step conditional distribution over future occupancy as its cumulant. This $k$-step conditional distribution $\mathbb{P}\left(s_{k}=\otherstate| s_0=s, \pi, \envprobs \right)$ describes the probability that the agent is at state $\otherstate$ after it follows policy $\pi$ for $k$ time-steps starting at state $s$. The SM is then defined as follows:
\begin{equation}\label{eq:sr-prob}
\horizon^\pi\left(\otherstate | s\right) 
  = (1-\gamma) \sum_{k=1}^\infty \gamma^k \mathbb{P}\left(s_{k+1}=\otherstate| s_0=s, \pi, \envprobs \right),
\end{equation}
where $(1-\gamma)$ ensures that $\horizon^\pi$ integrates to $1$. In the tabular setting, the SM is essentially a normalized SR, since
\begin{equation}\label{eq:sm-sr-relationship}
\horizon^\pi\left(\otherstate|s\right) = (1-\gamma) \sr^\pi\left(s, \otherstate\right).
\end{equation}
This relationship becomes apparent when we note that the expectation of an indicator function is the likelihood of the event, i.e. $\mathbb{E}[\mathbb{I}[X=x]] = \mathbb{P}(X=x)$. In the general (non-tabular) case, the SR and SM are not equivalent; this distinction has practical consequences for learning in realistic problems, where a tabular representation is not feasible.

Since the SM integrates to $1$, a key difference to the SR is that it defines a valid \textit{probability distribution}. 
This is important because it allows for the SR to generalize to continuous state and action spaces, where we can leverage density estimation techniques for estimating this value on a per-state basis.
As we will discuss in~\S\ref{sec:learning-horizon}, we can estimate the SM with density estimation techniques such as generative adversarial learning~\citet{janner2020gamma}, variational inference~\citep{thakoor2022generalised}, and contrastive learning~\citep{eysenbach2020c,zheng2023contrastive}.

SMs are interesting because they are a different kind of environment model. 
Rather than defining transition probabilities over next states, they describe the probability of reaching $\otherstate$ within a horizon determined by $\gamma$ when following policy $\pi$. 
While we don't know \textit{exactly} when $\otherstate$ will be reached, we can answer queries about \textit{whether} it will be reached within some relatively long time horizon with less computation compared to rolling out the base transition model.
Additionally, depending on how the SM is learned, we can also sample from it. This can be useful for policy evaluation and model-based control~\citep{thakoor2022generalised}. We discuss this in more detail in~\S\ref{sec:mbrl}.

Like the original SR, the SM obeys a Bellman-like recursion:
\begin{equation}\label{eq:horizon-bellman}
\horizon^\pi\left(\otherstate|s\right) = \rlexp{(1-\gamma) \envprobs(\otherstate|s, a) + \gamma \horizon^\pi\left(\otherstate|s'\right)},
\end{equation}
where the next-state probability $\otherstate$ in the first term resembles one-step reward in~\Eqref{eq:q-bellman} and the second term resembles the expected value function at the next time-step. 
As with~\Eqref{eq:sr-value}, we can use the SM to perform policy evaluation by computing:
\begin{align}\label{eq:horizon-value}
  V^{\pi}(s) 
    &= \frac{1}{1-\gamma} \mathbb{E}_{\otherstate \sim \horizon^{\pi}(\cdot|s)}\left[R(\otherstate)\right].
\end{align}
Additionally, we can introduce an action-conditioned variant of the SM
\begin{align}
  \horizon^\pi\left(\otherstate | s, a\right) 
  &= (1-\gamma) \sum_{t=0}^\infty \gamma^t \mathbb{P}\left(s_{t+1}=\otherstate| s_0=s, a_0=a, \pi, \envprobs \right) \nonumber \\
  &= (1-\gamma) \envprobs(\otherstate|s, a) + \gamma\rlexp{\horizon^\pi\left(\otherstate|s, a\right)}.
\end{align}
We can leverage this to compute an action value:
\begin{align}\label{eq:successor-policy-eval}
  Q^{\pi}(s, a) 
    &= \frac{1}{1-\gamma} \mathbb{E}_{\otherstate \sim \horizon^{\pi}(\cdot|s,a)}\left[R(\otherstate)\right].
\end{align}

\subsection{Successor features: a feature-based generalization of the SR}\label{sec:sf}

\begin{figure}
\centering
\includegraphics[width=\textwidth]{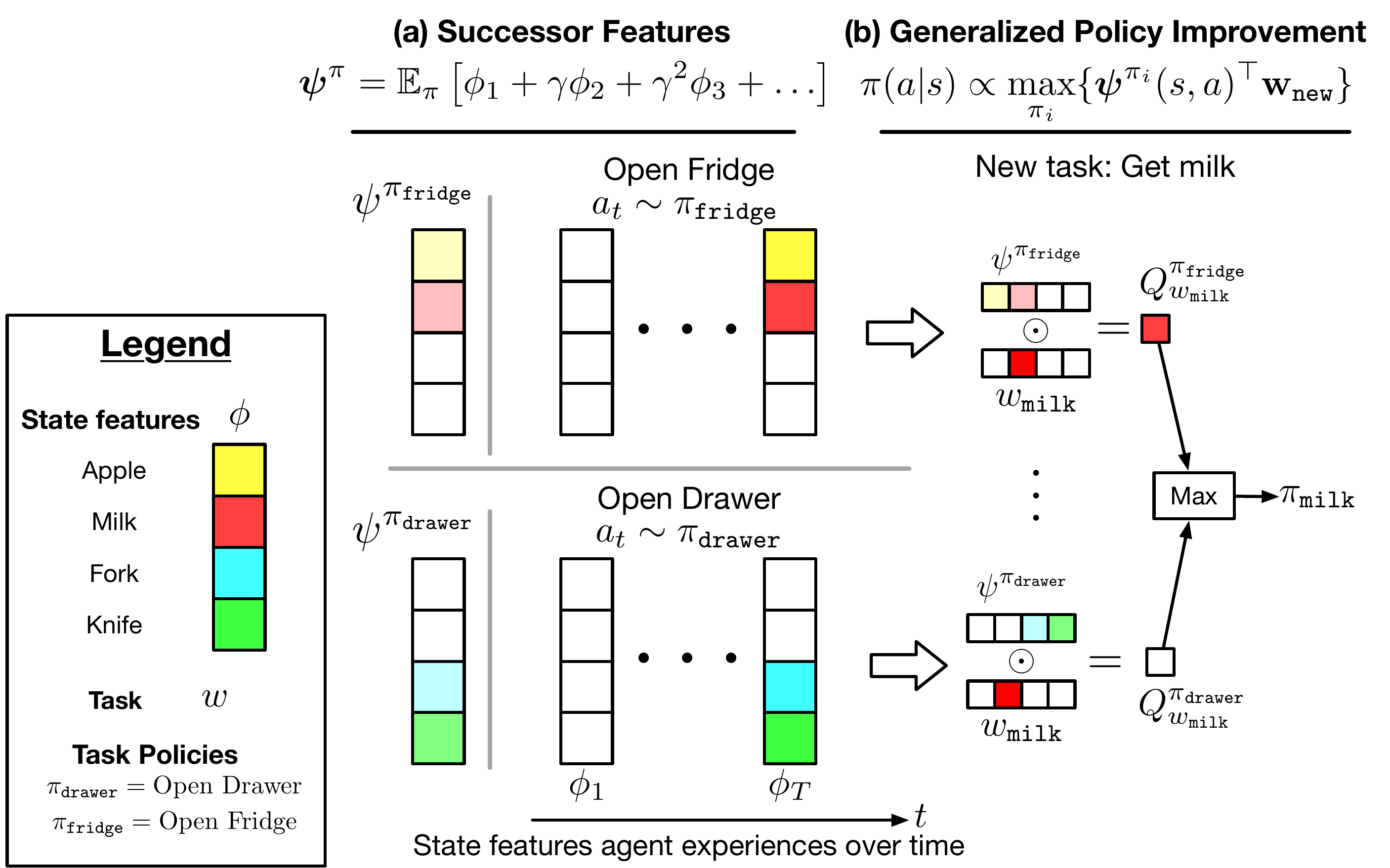}
\caption{A schematic of \textbf{Successor features (SFs; \S\ref{sec:sf})} and \textbf{Generalized policy improvement (GPI; \S\ref{sec:sf-gpi})}.
Note that we use the shorthand $\cumulant_t = \cumulant(s_t)$ to represents ``state-features'' that describe what is visible to the agent at time $t$. $\pi$ corresponds to policies that the agent knows how to perform. 
(a) Examples of SFs (\Eqref{eq:sf-def}) for the ``open drawer'' and ``open fridge'' policies. In this hypothetical scenario, the state-features that agent holds describe whether an apple, milk, fork, or knife are present.  Beginning for the first time-step, the SFs for these policies encode \textit{predictions} for which of these features will be present when the policies are executed---predicted to be present for apple and milk for the open fridge policy, and for the fork and knife when the open drawer policy is executed. 
(b) The agent can re-use these known policies with GPI (\Eqref{eq:sf-gpi}). When given a new task, say ``get milk'', it is able to leverage the SFs for the policies to knows to decide which behavior will enable it to get milk. In this example, the policy for opening the fridge will also lead to milk. The agent selects actions with GPI by computing Q-values for each known behavior as the dot-product between the current task and each known SF. The highest Q-value is then used to select actions. If the agent wants to execute the option keyboard (OK; \S\ref{sec:ok}), they can adaptively set $\task$ based on the current state. For example, at some states the agent may want to pursue getting milk, while in others they may want to pursue getting a fork. Adapted from \citep{carvalho2023combining} with permission.
}
\label{fig:sf-gpi-ok}
\end{figure}

When the state space is large or unknown, the SR can be challenging to compute or learn because it requires maintaining occupancy expectations over all states. This can also be challenging when dealing with learned state representations, as is common in practical learning algorithms (see \S\ref{sec:learning-sfs}). In these settings, rather than maintain occupancy measure for states, we can maintain occupancy measure over cumulants that are features shared across states, $\cumulant(s)$. This generalization of the SR is known as successor features~\citep[SFs;][]{barreto2017successor}. When $\cumulant\left(s\right)$ is a 1-hot vector describing the state the agent is in, SFs are exactly equivalent to the SR. 
The power of this representation is particularly apparent when reward can be decomposed into a dot-product of these successor features and a vector $\task$ describing feature preferences for the current task:
\begin{equation}\label{eq:sf-reward-relationship}
R(s,\mathbf{w})= \cumulant\left(s\right)^{\top} \task.
\end{equation}
SFs are then predictions of accumulated features $\cumulant$ the agent can expect to encounter when following a policy $\pi$:
\begin{equation}\label{eq:sf-def}
  \sf^\pi(s) = \rlexp{\sum_{t=0}^{\infty} \gamma^{t} \cumulant\left(s_{t+1}\right) \mid s_0=s}.
\end{equation}
Like the SR, SFs obey a Bellman equation:
\begin{equation}\label{eq:sf-bellman}
  \sf^\pi(s) = \rlexp{\cumulant(s') + \gamma \sf(s')}.
\end{equation}
Under the assumption of \Eqref{eq:sf-reward-relationship}, a task-dependent value $V^\pi(s, a, \task)$ is equivalent to:
\begin{align}\label{eq:sf-q-relationship}
  Q^\pi(s, a, \task) & =\rlexp{r(s_{1})+\gamma r(s_{2})+\ldots \mid s_0=s, a_0=a} \nonumber \\
  & =\rlexp{\cumulant(s_{1})^{\top} \task+\gamma \cumulant(s_{2})^{\top} \task+\ldots \mid s_0=s, a_0=a} \nonumber \\
  & =\rlexp{\sum_{t=0}^{\infty} \gamma^{t} \cumulant(s_{t+1}) \mid s_0=s, a_0=a}^{\top} \task \nonumber \\
  & = \sf^\pi(s)^{\top} \task .
\end{align}
As with the SR and SM, we can introduce an action-conditioned variant of SFs:
\begin{align}\label{eq:sf-def-2}
  \sf^\pi(s,a) &= \rlexp{\sum_{t=0}^{\infty} \gamma^{t} \cumulant\left(s_{t+1}\right) \mid s_0=s, a_0=a} \nonumber \\
  &= \rlexp{\cumulant(s') + \gamma \sf(s', a')}.
\end{align}
This provides an avenue to reuse these behavioral predictions for new tasks.
Once $\sf^\pi(s, a)$ has been learned, the behavior $\pi$ can be reused to pursue a novel reward function $R(s,\task_{\tt new})=\cumulant(s)^{\top}\task_{\tt new}$ by simply changing the corresponding task encoding $\task_{\tt new}$:
\begin{equation}\label{eq:sf-q-transfer}
  Q^\pi(s, a, \task_{\tt new}) = \sf^\pi(s, a)^{\top} \task_{\tt new}
\end{equation}
As we will see in the next subsection and when we discuss AI applications (\S\ref{sec:applications}), this becomes even more powerful when combined with an algorithm adaptively combining past policies. We discuss research on successor features over biologically plausible state features in~\S\ref{sec:bio-sr}.

\subsubsection{Generalized policy improvement: adaptively combining policies}\label{sec:sf-gpi}
One limitation of~\Eqref{eq:sr-value} and~\Eqref{eq:horizon-value} is that they only enable us to re-compute the value of states for a new reward function under a known policy. However, we may want to \textit{synthesize} a new policy from the other policies we have learned so far. 
We can accomplish this with SFs by combining them with Generalized Policy improvement ~\citep[GPI;][]{barreto2017successor}, illustrated in Figure \ref{fig:sf-gpi-ok}.

Assume we have learned (potentially optimal) policies $\left\{\pi_i\right\}_{i=1}^{n_{\text {train}}}$ and their corresponding SFs $\left\{\sf^{\pi_i}(s, a)\right\}_{i=1}^{n_{\text {train}}}$ for $n_{\tt train}$ training tasks $\left\{\task_i\right\}_{i=1}^{n_{\text {train}}}$. When presented with a new task $\task_{\tt new}$, we can obtain a new policy with GPI in two steps: (1) compute Q-values using the training task SFs; (2) select actions using the highest Q-value. This operation is summarized as follows:
\begin{align}\label{eq:sf-gpi}
  a^\ast\left(s; \task_{\tt new}\right)
  &=
  \argmax_a \max _{i \in\left\{1, \ldots, n_{\text {train}}\right\}}
    \left\{\sf^{\pi_i}(s, a)^{\top} \task_{\tt new}\right\} \nonumber \\
  &=\argmax_a
  \max _{i \in\left\{1, \ldots, n_{\text {train}}\right\}}
    \left\{Q^{\pi_i}(s,a, \task_{\tt new})\right\}.
\end{align}
If $\task_{\tt new}$ is in the span of the training tasks (i.e., $\task_{\tt new} = \sum_{i} \alpha_i \task_i$, where $\alpha_i \in \mathbb{R}$), the GPI theorem states that $\pi(a|s) = \mathbb{I}[a=a^\ast(s,\task_{\tt new})]$ will perform at least as well as any of the training policies---i.e., $Q^{\pi}(s,a, \task_{\tt new}) \geq \max _i Q^{\pi_i}(s,a, \task_{\tt new})\, \forall(s, a) \in(\mathcal{S} \times \mathcal{A})$. We discuss how this has been used for transferring policies in artificial intelligence in~\S\ref{sec:ai-transfer}. We discuss empirical evidence that humans transfer policies with GPI in~\S\ref{sec:MTRL}.

\subsubsection{Option Keyboard: chaining together policies}\label{sec:ok}
One advantage of ~\Eqref{eq:sf-gpi} is that it facilitates adaptation to linear combinations of training task encodings. However, when transferring to a new task encoding $\task_{\tt new}$, the preferences are constant over time. This becomes problematic when dealing with complex tasks that necessitate different preferences for different states---for example, tasks that require both avoidance and approach behaviors at different times. 

The \emph{Option Keyboard} \citep{barreto2019option,barreto2020fast} introduces a solution to this by employing a \textit{state-dependent} preference vector $\task_s$. This vector is generated through a policy function $g(\cdot)$, which takes as input the current state $s$ and the new task $\task_{\tt new}$, $\task_s = g(s, \task_{\tt new})$. Actions can then be chosen as follows:
\begin{align} \label{eq:ok}
  a^\ast_{\task_s}\left(s; \task_{\tt new}\right) = \argmax_a \max _{i \in \{1, \ldots, n_{\tt train}\} }\left\{\sf^{\pi_{i}}(s,a)^{\top} \task_s\right\} 
\end{align}
Note that this policy is \textit{adaptive}---which known policy $\pi_i$ is chosen at a time-step is dependent on which one maximizes the Q-values at that time-step. This is why it's called the Option Keyboard: $\task_s$ will induce a set of policies to be active for a period of time, somewhat like playing a chord on a piano. We discuss how this has been in artificial intelligence applications to chain policies in~\S\ref{sec:ok-transfer}.

\subsection{Summary}

The predictive representations introduced above can be concisely organized in terms of particular cumulants, as summarized in Table \ref{table:sr-summary}. These cumulants have different strengths and weaknesses. Value functions (reward cumulants) directly represent the key quantity for RL tasks, but they suffer from poor flexibility. The SR (state occupancy cumulant) and its variations (feature occupancy and state probability cumulants) can be used to compute values, but also retain useful information for generalization to new tasks (e.g., using generalized policy improvement). 

\begin{table}[ht]
  \centering
  \begin{tabular}{|l|l|l|}
  \hline
  Predictive representation & Cumulant & $\delta$: TD update when $a' \sim \pi$ and $s' \sim T$ \\
  \hline
  $Q^{\pi}(s,a)$ (\S\ref{sec:algorithmic}) & $r(s)$ & $r(s') + \gamma Q^{\pi}(s',a') - Q^{\pi}(s,a)$ \\
  \hline
  $\sr^{\pi}(s,\otherstate)$ (SR; \S\ref{sec:sr}) & $\mathbb{I}(s=\otherstate)$ & $\mathbb{I}(s'=\otherstate) + \gamma \sr^{\pi}(s',\otherstate) - \sr^{\pi}(s,\otherstate)$ \\
  \hline
  $\sf^{\pi}(s,a)$ (SFs; \S\ref{sec:sf}) & $\cumulant(s)$ & $\cumulant(s') + \gamma \sf^{\pi}(s',a') - \sf^{\pi}(s,a)$ \\
  \hline
  $\horizon^{\pi}(\otherstate|s,a)$ (SM; \S\ref{sec:successor-models}) & $\mathbb{P}\left(s_{t+1}=\otherstate| s_0, a_0 \right)$ & $(1-\gamma)T(s'|s,a) + \gamma \horizon^{\pi}(\otherstate|s',a') - \horizon^{\pi}(\otherstate|s,a)$ \\
  \hline
  \end{tabular}
  \caption{\textbf{Summary of predictive representations that we focus on}. For each, we also describe the ``cumulant'' that this predictive representation forms predictions over, along with a corresponding on-policy Bellman update one can use to learn the representation for a policy $\pi$. 
  $Q^{\pi}(s,a)$ is the action-value function which forms predictions about future reward.
  $\sr^{\pi}(s,\otherstate)$ is the successor representation (SR) which forms predictions about how much a state $\otherstate$ will be visited. 
  $\sf^{\pi}(s,a)$ are successor features (SFs) which form predictions about how much state-features $\cumulant(s)$ will be experienced.
  $\horizon^{\pi}(\otherstate|s,a)$ is the successor model (SM) which predicts the likelihood of experiencing $\otherstate$ in the future.
  }
  \label{table:sr-summary}
  \end{table}

\section{Practical learning algorithms and associated challenges}\label{sec:learning}

Of the predictive representations discussed in \S\ref{sec:theory}, only SFs and SMs have been successfully scaled to environments with large, high-dimensional state spaces (including continuous state spaces). Thus, these will be our focus of discussion.
We will first discuss learning SFs in \S\ref{sec:learning-sfs} and then learning successor models in \S\ref{sec:learning-horizon}. 

\subsection{Learning successor features}\label{sec:learning-sfs}

In this section we discuss practical considerations for learning successor features, including learning a function $\cumulant_{\theta}$ that produces cumulants (\S\ref{sec:learning-cumulants}) and estimating SFs $\sf_{\theta}$ (\S\ref{sec:estimating-sfs}).

\subsubsection{Discovering cumulants}\label{sec:learning-cumulants}

One central challenge to learning SFs is that they require cumulants that are useful for adaptive agent behavior; these are not always easy to define \emph{a priori}. In most work, these cumulants have been hand-designed, but potentially better ones can be learned from experience. Some general methods for discovering cumulants include leveraging meta-gradients~\citep{veeriah2019discovery}, discovering features that enable reconstruction~\citep{kulkarni2016deep,machado2017eigenoption}, and maximizing the mutual information between task encodings and the cumulants that an agent experiences when pursuing that task~\citep{hansen2019fast}. However, these methods don't necessarily guarantee that the learned cumulants respect a linear relationship with reward (\Eqref{eq:sf-reward-relationship}). To satisfy this, methods typically enforce this by minimizing the L2-norm of their difference~\citep{barreto2017successor}:
\begin{equation}\label{eq:reward-loss}
  \mathcal{L}_r = || r - \cumulant_{\theta}(s)^{\top} \task ||^2_2.
\end{equation}
When learning cumulants that support transfer with GPI, one strategy that can bolster~\Eqref{eq:reward-loss} is to learn an $n$-dimensional $\cumulant$ vector for $n$ tasks such that each dimension predicts one task reward~\citep{barreto2018transfer}. Another strategy is to enforce that cumulants describe independent features~\citep{alver2021constructing}---e.g., by leveraging a modular architecture with separate parameters for every cumulant dimension~\citep{carvalho2023composing} or by enforcing sparsity in the cumulant dimensions~\citep{filos2021psiphi}. We discuss research on biologically plausible state features in~\S\ref{sec:bio-sr}.

\subsubsection{Estimating successor features}\label{sec:estimating-sfs}

\paragraph{Learning an estimator that can generalize across policies}
The first challenge for learning SFs is that they are defined for a particular policy $\pi$. We can mitigate this by learning \textit{Universal Successor Feature Approximators}~\citep[USFAs;][]{borsa2018universal}, which takes as input a  \textit{policy encodings} $\mathbf{z}_{\task} \in \mathcal{Z}_{\pi}$, which represent an encoding of the policy $\pi_{\task}$ that is reward-maximizing for task $\task$. Concretely, we can estimate $\sf^{\pi_\task}(s,a)$ as:
\begin{equation}
  \sf^{\pi_\task}(s,a) \approx \sf_{\theta}(s,a, \mathbf{z}_{\task}).
\end{equation}
This has several benefits. First, one can share the estimator parameters across policies, which can improve learning.
Second, this allows SFs to generalize across different policies.
Leveraging a USFA, the GPI operation in~\Eqref{eq:sf-gpi} can be adapted to perform a max operation over $\mathcal{Z}_{\pi}$:
\begin{align}\label{eq:usfa-gpi}
  a^\ast\left(s; \task_{\tt new}\right)
  &=
  \underset{a \in \mathcal{A}}{\arg \max } \max _{\mathbf{z}_{\task} \in \mathcal{Z}_{\pi}}
  \left\{\sf_{\theta}(s, a, \mathbf{z}_{\task})^{\top} \task_{\tt new}\right\}
\end{align}
Several algorithms exploit this property (see \S\ref{sec:no-reward}, \S\ref{sec:non-task}, \S\ref{sec:mulit-agent}).
If the policy encoding space is equivalent to the $n$ training tasks, $\mathcal{Z}_{\pi}=\{\task_{1}, \ldots, \task_n\}$, this recovers the original GPI operation (\Eqref{eq:sf-gpi}). In practice, it is common to identify the policy encoding $\mathbf{z}_{\task}$ for a task with its  task encoding $\task$, i.e. to use $\mathbf{z}_{\task}=\task$. %

\paragraph{Learning successor features while learning a policy} Often SFs need to be learned simultaneously with policy optimization. This requires the SFs to be updated along with the policy. One strategy is to simultaneously learn an action-value function $Q_{\theta}(s,a,\task)=\sf_{\theta}(s,a,\mathbf{z}_{\task})^{\top}\task$ that is used to select actions. One can accomplish this with Q-learning over values defined by the SFs and task encoding. Q-values are updated towards a target $\mathbf{y}_Q$ defined as the sum of the reward $R(s'; \task)$ and the best next Q-value. We define the learning objective $\mathcal{L}_Q$ as follows:
\begin{align}\label{eq:sf-q-loss}
  \mathbf{y}_Q &= R(s'; \task) + \gamma \sf_{\theta}(s',a^*,\mathbf{z}_{\task})^{\top}\task
    & \mathcal{L}_Q &= || \sf_{\theta}(s,a,\mathbf{z}_{\task})^{\top}\task - \mathbf{y}_Q ||
\end{align}
where $a^* = \argmax_{a'} \sf_{\theta}(s',a',\mathbf{z}_{\task})^{\top}\task$ is the action which maximizes features determined by $\task$ at the next time-step.
To ensure that these Q-values follow the structure imposed by SFs (\Eqref{eq:sf-q-relationship}), we additionally update SFs at a state with a target $\mathbf{y}_{\sf}$ defined as the sum of the cumulant $\cumulant(s')$ and the SFs associated with best Q-value at the next state:
\begin{align}\label{eq:sf-scalar-loss}
  \mathbf{y}_{\sf} &= \cumulant(s') + \gamma \sf_{\theta}(s',a^*,\mathbf{z}_{\task}) 
    & \mathcal{L}_{\sf} &=  || \sf_{\theta}(s,a,\mathbf{z}_{\task}) - \mathbf{y}_{\sf}||.
\end{align}
In an online setting, it is important to learn SFs with data collected by a policy which chooses actions with high Q-values. This is especially important if the true value is lower than the estimated Q-value. Because Q-learning leverages the \textit{maximum} Q-value when doing backups, it has a bias for over-estimating value. This can destabilize learning, particularly in regions of the state space that have been less explored~\citep{ostrovski2021difficulty}. 

Another strategy to stabilize SF learning is to learn individual SF dimensions with separate modules~\citep{carvalho2023composing,carvalho2023combining}. Beyond stabilizing learning, this modularity also enables approximating SFs that generalize better to novel environment configurations (i.e., which are more robust to novel environment dynamics).

\paragraph{Estimating successor features with changing cumulants} In some cases, the cumulant itself will change over time (e.g., when the environment is non-stationary). This is challenging for SF learning because the prediction target is non-stationary~\citep{barreto2018transfer}. This is an issue even when the environment is stationary but the policy is changing over time: different policies induce different trajectories and different state-features induce different descriptions of those trajectories. %

One technique that has been proposed to facilitate modeling a non-stationary cumulant trajectories is to learn an SF as a probability mass function (pmf) $p(\sf^{(k)}|s,a,\task)$ defined over some set of possible values $\mathbb{B}=\left\{b_1, \ldots, b_M\right\}$. Specifically, we can estimate an n-dimensional SF $\sf^{\pi_\task}(s, a) \in \mathbb{R}^n$ with $\sf_\theta(s, a, \mathbf{z}_{\task})$ and represent the $k$-th SF dimension as $\sf_\theta^k(s, a, \mathbf{z}_{\task})=$ $\sum_{m=1}^M p(\sf^{(k)}=b_m|s,a,\mathbf{z}_{\task}) b_m$. We can then learn SFs with a negative log-likelihood loss where we construct categorical target labels $\mathbf{y}_{\sf^{(k)}}$ from the return associated with the optimal Q-value~\citep{carvalho2023combining}:
\begin{align}\label{eq:sf-pmf-loss}
  \mathbf{y}_{\sf^{(k)}} &= \cumulant_{\theta}^k(s') + \gamma \sf^k_{\theta}(s',a^*,\mathbf{z}_{\task}) \\
     \mathcal{L}_{\sf} &=  - \sum^n_{k=1} f_{\tt label}(\mathbf{y}_{\sf^{(k)}})^{\top} \log p(\sf^{(k)}|s,a,\mathbf{z}_{\task}).
\end{align}
Prior work has found that the two-hot representation is a good method for defining $f_{\tt label}(\cdot)$ ~\citep{carvalho2023combining,schrittwieser2020mastering}. In general, estimating predictive representations such as SFs with distributional losses such as \Eqref{eq:sf-pmf-loss} has been shown to reduce the variance in learning updates~\citep{imani2018improving}. This is particularly important when cumulants are being learned, as this can lead to high variance in $\mathbf{y}_{\sf^{(k)}}$.

\subsection{Learning successor models}\label{sec:learning-horizon}
In this section, we focus on estimating $\horizon^{\pi}(\otherstate|s,a)$ with $\horizon_{\theta}(\otherstate|s,a)$.
In a tabular setting, one can leverage TD-learning with the Bellman equation in~\Eqref{eq:horizon-bellman}. However, for very large state spaces (such as with infinite size continuous state spaces), this is intractable or impractical. Depending on one's use-case, different options exist for learning.
First, we discuss the setting where one wants to learn an SM they can sample from (\S\ref{sec:learning-horizon-sample}).
Then we discuss the setting where one only wants to \textit{evaluate} an SM for different actions given a target state (\S\ref{sec:learning-horizon-eval}).

\subsubsection{Learning successor models that one can sample from}\label{sec:learning-horizon-sample}
Learning an SM that can be sampled from is useful for evaluating a policy (\Eqref{eq:horizon-value}), evaluating a sequence of policies~\citep{thakoor2022generalised}, and in model-based control~\citep{janner2020gamma}.
There are two ways to learn such SMs: adversarial learning and density estimation~\citep{janner2020gamma}. Adversarial learning has been found to be unstable, so we focus on density estimation, where the objective is to find parameters $\theta$ that maximize the log-likelihood of states sampled from $\horizon^\pi$:
\begin{align}\label{eq:horizon-learning-density}
  \mathcal{L}_{\mu} = \mathbb{E}_{\substack{
    (s,a) \sim p(s,a) \\
    \otherstate \sim \horizon^{\pi}(\cdot \mid s, a)}}\left[ \log \horizon_\theta\left(\otherstate \mid s, a\right) \right]
\end{align}
We can optimize this objective as follows. When sampling targets, we need to sample $\otherstate$ in proportion to the discount factor $\gamma$. We can accomplish this by first sampling a time-step from a geometric distribution, $k\sim\operatorname{Geom}(1-\gamma)$, and then selecting the state at that time-step, $s_{t+k}$, as the target $\tilde{s}$. 

While this is a simple strategy, it has several challenges. For values of $\gamma$ close to 1, this becomes a challenging learning problem requiring predictions over very long time horizons. Another challenge is that you are using $s_{t+k}$ obtained under policy $\pi$. In practice, we may want to leverage data collected under a different policy. This happens when, for example, we want to learn from a collection of different datasets, or we are updating our policy over the course of learning. Learning from such off-policy data can lead to high bias, or a high variance learning update from off-policy corrections~\citep{precup2000eligibility}.

We can circumvent these challenges as follows. First let's define a Bellman operator $T^\pi$:
\begin{equation}
    (T^{\pi} \horizon^{\pi})(\otherstate|s,a) = (1-\gamma) T(s'|s,a) + \gamma \sum_{s'} T(s'|s,a) \sum_{a'}\pi(a'|s')\horizon(\otherstate|s',a').
\end{equation}
With this we can define a cross-entropy temporal-difference loss~\citep{janner2020gamma}:
\begin{equation}\label{eq:cetd}
    \mathcal{L}_{\mu} = \mathbb{E}_{\substack{
    (s,a) \sim p(s,a) \\
    \otherstate \sim (T^{\pi} \horizon^{\pi})(\cdot|s,a)}}\left[ \log \horizon_\theta\left(\otherstate \mid s, a\right) \right]
\end{equation}
Intuitively, $(T^{\pi} \horizon^{\pi})(\cdot|s,a)$ defines a random variable obtained as follows. First sample $s'\sim T(\cdot|s,a)$. Terminate and emit $s'$ with probability $(1-\gamma)$. Otherwise, sample $a'\sim\pi(a'|s')$ and them sample $\otherstate \sim \horizon^{\pi}(\cdot |s',a')$.

The most recent promising method for learning \eqref{eq:cetd} has been to leverage a variational autoencoder ~\citep{thakoor2022generalised}. Specifically, we can define an approximate posterior $q_{\psi}(z|s,a,\otherstate)$ and then optimize the following evidence lower-bound:
\begin{equation}
    \mathcal{L}_{\mu} = \mathbb{E}_{\substack{
    (s,a) \sim p(s,a) \\
    \otherstate \sim (T^{\pi} \horizon^{\pi})(\cdot|s,a) \\ z \sim q_{\psi}(\cdot|s,a,\otherstate)}} \left[
            \log \frac{\horizon_\theta\left(\otherstate \mid s, a, z \right) } {q_{\psi}(z|s,a,\otherstate)}
    \right].
\end{equation}
While \citet{thakoor2022generalised} were able to scale their experiments to slightly more complex domains than~\citet{janner2020gamma}, their focus was on composing policies via Geometric Policy Composition (discussed more in \S\ref{sec:transfer-advances}), so it is unclear how well their method performs in more complex domains. 
The key challenge for this line of work is in sampling from $\horizon^{\pi}(\cdot \mid s, a)$, where $\otherstate$ can come from a variable next time-step after $(s,a)$. In the next section, we discuss methods that address this challenge. 

\subsubsection{Learning successor models that one can evaluate}\label{sec:learning-horizon-eval}
Sometimes we may not need to learn an SM that we can sample from, only one that we can evaluate. This can be used, for example, to generate and improve a policy that achieves some target state~\citep{eysenbach2020c,zheng2023contrastive}.
One strategy is to learn a classifier $p^{\mu}_{\theta}$ that, given $(s,a)$, computes how likely $\otherstate$ is compared to some set of $N$ random (negative) states $\{s^{-}_{i}\}^N_{i=1}$ the agent has experienced:
\begin{equation}\label{eq:horizon-classifier}
  p^{\mu}_{\theta}(\otherstate \mid s,a, s^{-}_{1:N})
  =
  \frac{\exp(f_{\theta}(s,a, \otherstate))}{
    \exp(f_{\theta}(s,a, \otherstate)) + \sum^N_{i=1} \exp(f_{\theta}(s,a, s^{-}_{i}))
    }.
\end{equation}
The function $f_\theta$ can be chosen to maximizes classification accuracy across random states and actions in the agent's experience, $(s,a) \sim p(s,a)$, target states $\otherstate$ drawn from the empirical successor model distribution, $\otherstate \sim {\horizon^{\pi}(\otherstate \mid s,a)}$, and negatives drawn from the state-marginal, $s^{-}_{1:N} \sim p(s)$:
\begin{equation}\label{eq:successor-ratio}
  L_{\horizon} = \mathbb{E}_{\substack{
    (s,a) \sim p(s,a) \\
    \otherstate \sim {\horizon^{\pi}(\otherstate \mid s,a)} \\
    s^{-}_{1:N} \sim p(s)
  }}\left[
    \log p^{\mu}_{\theta}(\otherstate \mid s,a, s^{-}_{1:N})
  \right].
\end{equation}
If we can find such an $f_{\theta}$, then the resulting classifier is approximately equal to the density ratio between $\horizon^{\pi}(\otherstate \mid s,a)$ and the state marginal $\otherstatemarginal(\otherstate)$~\citep{poole2019variational,zheng2023contrastive}:
\begin{equation}\label{eq:horizon-ratio}
  \frac{\horizon^{\pi}(\otherstate \mid s,a)}{\otherstatemarginal(\otherstate)} 
  \approx
  (N+1) \cdot p^{\mu}_{\theta}(\otherstate \mid s,a, s^{-}_{1:N}).
\end{equation}
Optimizing~\Eqref{eq:successor-ratio} is challenging because it requires sampling from $\horizon^{\pi}$. 
We can circumvent this by instead learning the following TD-like objective, where we replace sampling from $\horizon^{\pi}$ with sampling from the state-marginal and reuse $p_{\theta}^{\mu}$ as an importance weight:
\begin{equation}\label{eq:td-horizon-ratio}
  L_{\horizon} = \mathbb{E}_{\substack{
    (s,a) \sim p(s,a) \\
    s' \sim \envprobs (\cdot |s,a) \\
    a' \sim \pi(a|s') \\
    \tilde{s} \sim p(s) \\
    s^{-}_{1:N} \sim p(s)
  }}\left[
      (1- \gamma) \log p^{\mu}_{\theta}(s'|s,a, s^{-}_{1:N})
      + \gamma 
        (N+1)p_{\theta}^{\mu}(\otherstate|s',a',s^{-}_{1:N})  \log p_{\theta}^{\mu}(\otherstate|s,a,s^{-}_{1:N})
  \right].
\end{equation}
\citet{zheng2023contrastive} show that (under some assumptions) optimizing~\Eqref{eq:td-horizon-ratio} leads to the following Bellman-like update:
\begin{equation}\label{eq:horizon-ratio-bellman}
  p_{\theta}^{\mu}(\otherstate|s,a,s^{-}_{1:N}) \leftarrow (1-\gamma)\envprobs(s'=\otherstate|s,a) + \gamma \mathbb{E}_{\substack{s' \sim T (\cdot|s,a) \\a' \sim \pi(\cdot|s')}} \left[p_{\theta}^{\mu}(\otherstate|s',a',s^{-}_{1:N})\right],
\end{equation}
which resembles the original SM Bellman equation (\Eqref{eq:horizon-bellman}). 
However, one key difference to~\Eqref{eq:horizon-bellman} is that we parametrize $p_{\theta}^{\mu}(\otherstate|s,a,s^{-}_{1:N})$ with $N+1$ random samples (e.g.,~from a replay mechanism), which is a form of contrastive learning.
This provides a nonparametric algorithm for learning SMs. 
With the SR, one performs the TD update for all states (\Eqref{eq:SRTDerror}); here, one performs this update using a random sample of $N+1$ states. 

We can define $f_{\theta}$ as the dot product between a predictive representation $\varphi_{\theta}(\cdot,\cdot)$ and label representation $\phi_{\theta}(\cdot)$, $f_{\theta}(s,a, \otherstate) = \varphi_{\theta}(s,a)^{\top}\phi_{\theta}(\otherstate)$.
{$\phi_{\theta}(s)$ can then be thought of as state-features analogous to SFs (\S\ref{sec:sf}). $\varphi_{\theta}(s,a)$ is a prediction of these future features similar to SFs, $\sf_{\theta}(s,a)$, with labels coming from future states; however, it doesn't necessarily have the same semantics as a discounted sum (i.e., \Eqref{eq:sf-q-relationship}). We use similar notation because of their conceptual similarity.} 
We can then understand~\Eqref{eq:td-horizon-ratio} as doing the following. 
The first term in this objective pushes the prediction $\varphi_{\theta}(s,a)$ towards the features at the next-timestep $\phi_{\theta}(s')$, and the second term pushes $\varphi_{\theta}(s,a)$ towards the features at arbitrary state-features $\phi_{\theta}(\otherstate)$. Both terms repel $\varphi_{\theta}(s,a)$ from arbitrary ``negative'' state-features $\phi_{\theta}(\otherstate^-_{i})$.

\section{Artificial intelligence applications}

\label{sec:applications}

In this section, we discuss how the SR and its generalizations have enabled advances in artificial agents that learn and transfer policies.

\subsection{Exploration}\label{sec:exploration}

\subsubsection{Pure exploration}\label{sec:no-reward}
\paragraph{Learning to explore and act in the environment before exposure to reward.} In the ``pure exploration'' setting, an agent can explore its environment for some period of time without external reward. In some cases, the goal is to learn a policy that can transfer to an unknown task $\task_{\tt new}$. SFs can be used to achieve such transfer.

One useful property of SFs is that they encode predictions about what features one can expect when following a policy. Before reward is provided, this can be used to reach different parts of the state space with different policies. One strategy is to associate different parts of the state space with different parts of a high-dimensional task embedding space~\citep{hansen2019fast}. At the beginning of each episode, an agent samples a goal encoding $\task$ from a high-entropy task distribution $p(\task)$. 
During the episode, the agent selects actions that maximize the features described by $\task$ (e.g., with~\Eqref{eq:sf-q-relationship}).
As it does this, it learns to predict $\task$ from the states it encounters. If $p(\mathbf{w})$ is parametrized as a von Mises distribution, then the agent can learn this prediction by simply maximizing the dot-product between the state-features $\cumulant(s)$ and goal encoding $\task$:
\begin{align}\label{eq:loss-visr}
  \mathcal{L}_{\cumulant} = \log p_{\theta}(\task|s) = \cumulant_{\theta}(s)^{\top} \task.
\end{align}
This is equivalent to maximizing the mutual information between $\cumulant_{\theta}(s)$ and $\task$.
As the agent learns cumulants, it learns SFs as usual (e.g., with~\Eqref{eq:sf-scalar-loss}).
Thus, one can essentially use a standard SF learning algorithm and simply replace the cumulant discovery loss (e.g.,~\Eqref{eq:reward-loss}) with~\Eqref{eq:loss-visr}.
Once the agent is exposed to task reward $\task_{\tt new}$, it can freeze $\cumulant_{\theta}$ and solve for $\task_{\tt new}$ (e.g., with~\Eqref{eq:reward-loss}).
The agent can then use GPI to find the best policy for achieving $\task_{\tt new}$ by searching over a Gaussian ball defined around $\task_{\tt new}$. This is equivalent to setting $Z=\{z_i \sim \mathcal{N}(\task_{\tt new}, \sigma)\}^n_{i=1}$ for~\Eqref{eq:usfa-gpi}, where $\sigma$ defines the size of the ball. 

~\citet{hansen2019fast} leveraged this strategy to develop agents which could explore Atari games without any reward for 250 million time-steps and then have 100 thousand time-steps to earn reward. They showed that this strategy was able to achieve superhuman performance across most Atari games, despite not observing any rewards for most of its experience. 
\citet{liu2021aps} improved on this algorithm by adding an intrinsic reward function that favors exploring parts of the state space that are surprising (i.e., which induce high entropy) given a memory of the agent's experience. This dramatically improved sample efficiency for many Atari games.

\paragraph{Exploring the environment by building a map.} One strategy for exploring large state spaces systematically is to build a map of the environment defined over landmarks in the environment.
With such a map, an agent can systematically explore by planning paths towards the frontier of its knowledge~\citep{ramesh2019successor,hoang2021successor}. 
Numerous questions arise in this process. How do we define good landmarks? How do we define policies that traverse between landmarks? How does an agent identify that it has made progress between landmarks after it has set a plan, or course-correct if it finds that it accidentally deviated?~\cite{hoang2021successor} developed an elegant solution to all of these problems with the \textit{successor feature similarity} (SFS) metric.
This similarity metric defines closeness of two states by the frequency with which an agent starting in each state visits the same parts of the environment. Concretely:
\begin{align}\label{eq:sfsim}
  S_{\sf}(s_1,a, s_2) &= \sf^{\bar{\pi}}(s_1,a)^{\top} \mathbb{E}_{A\sim\bar{\pi}}[\sf^{\bar{\pi}}(s_2,A)] \\
  S_{\sf}(s_1,s_2) &= \mathbb{E}_{A\sim\bar{\pi}}[\sf^{\bar{\pi}}(s_1,a)^{\top}] \mathbb{E}_{A\sim\bar{\pi}}[\sf^{\bar{\pi}}(s_2,A)],
\end{align}
where $\bar{\pi}$ is a uniform policy.\footnote{Note that to avoid excessive notation, we've overloaded the definition of $S_\psi$.}

Through \textit{only} learning of SFs over pretrained cumulants $\cumulant$,~\citet{hoang2021successor} were able to address all the needs above by exploiting SFS.
Let $\mathcal{L} \subset \mathcal{S}$ be the set of landmarks discovered so far. The algorithm works as follows. Each landmark $L \in \mathcal{L}$ has associated with it a count $N(L)$ for how often its been visited. A ``frontier'' landmark $L_F$ is sampled in proportion to the inverse of this count. The agent makes a shortest-path plan of subgoals $(s^\ast_1, \ldots, s^\ast_n)$ towards this landmark where $s_n=L_F$. In order to navigate to the next subgoal $s^\ast_i$, it defines a policy with the action of the current successor features that are most aligned with the goal's successor features, $a^\ast(s, s^\ast_i) = \argmax_a S_{\sf}(s,a,s^\ast_i)$. As it traverses towards the landmark, it localizes itself by comparing the current state $s$ to known landmarks $f_{\tt loc}(s, \mathcal{L}) = \argmax_{L \in \mathcal{L}} S_{\sf}(s, L)$. When $s^\ast_i = f_{\tt loc}(s, \mathcal{L})$, it has reached the next landmark and it moves on to the next subgoal. 
Once a frontier landmark is found, the agent explores with random actions. The states it experiences are added to its replay buffer and used in $2$ ways. First, they update the agent's current SR. Second, if a state is sufficiently different from known landmarks, it is added. 
In summary, the $3$ key functions that one can compute without additional learning are:
\begin{align*}
  \pi(s, s^\ast_i) &= \argmax_a S_{\sf}(s,a,s^\ast_i) && \text{goal-conditioned policy} \\
  f_{\tt loc}(s, \mathcal{L}) &= \argmax_{L \in \mathcal{L}} S_{\sf}(s, L) && \text{localization function} \\
  f_{\tt add}(s, \mathcal{L}) &= \mathcal{L} \leftarrow \mathcal{L} \cup s \quad \text{if}\quad (S_{\sf}(s, L) < \epsilon_{\tt add} )\,\forall L \in \mathcal{L} && \text{landmark addition function}
\end{align*}
The process then repeats. This exploration strategy was effective in exploring both minigrid~\citep{MinigridMiniworld23} and the partially observable 3D VizDoom environments~\citep{kempka2016vizdoom}.

\subsubsection{Balancing exploration and exploitation}\label{sec:exploration-reward}
\paragraph{Cheap uncertainty computations.}
Balancing exploration and exploitation is a central challenge in RL~\citep{sutton18, kaelbling1996reinforcement}. One method that provides close to optimal performance in tabular domains is posterior sampling \citep[also known as Thompson sampling, after][]{thompson1933likelihood}, where an agent updates a posterior distribution over Q-values and then chooses the value-maximizing action for a random sample from this distribution \citep[see][for a review]{russo2018tutorial}. The main difficulties for implementing posterior sampling are associated with representing, updating, and sampling from the posterior when the state space is large. \citet{janz2019successor} showed that SFs enable cheap method for posterior sampling.  They assume the following prior, likelihood, and posterior for the environment reward:
\begin{align}
  \task \sim \mathcal{N}(0, \mathbf{I}) \quad\quad\quad
  r|\task \sim \mathcal{N}(\cumulant(s)^\top \task, \beta) \quad\quad\quad
  \task|r,s \sim \mathcal{N}(\horizon_{\task}, \Sigma_{\task})
\end{align}
where $\beta$ is the variance of reward around a mean that is linear in the state feature $\cumulant(s)$; $\horizon_{\task}$ and $\Sigma_{\task}$ are given by analytical solutions for the mean and variance of the posterior given the Gaussian prior/likelihood assumptions and a set of observations. The posterior distribution for the Q-values is then given by:
\begin{equation}
  \hat{Q}_{\mathrm{SU}}^\pi \sim \mathcal{N}\left({\Psi}^\pi \horizon_{\task}, {\Psi}^\pi \Sigma_{\task}\left({\Psi}^\pi\right)^{\top}\right)
\end{equation}
where ${\Psi}^\pi=[\sf^{\pi}(s,a)]^{\top}_{(s,a) \in \mathcal{S} \times \mathcal{A}}$ is a matrix with each row corresponding to an SF for a state-action pair. SFs, cumulants, and task encodings are learned with standard losses (e.g.,~\Eqref{eq:reward-loss} and~\Eqref{eq:sf-scalar-loss}).

\paragraph{Count-based exploration.} Another method that provides (near) optimal exploration in tabular settings is count-based exploration with a bonus of $1/\sqrt{N(s)}$~\citep{auer2002using}, where $N(s)$ is the number of times a state $s$ has been visited. When the state space is large, it can be challenging to track this count.~\citet{machado2020count} showed that that the L1-norm of SFs is proportional to visitation count, and can therefore be used as an exploration bonus:
\begin{equation}
  R^{\tt int}(s) = \frac{1}{||\sf^{\pi}(s)||_1}.
\end{equation}
With this exploration bonus, they were able to improve exploration in sparse-reward Atari games such as Montezuma's revenge. Recent work has built on this idea: \citet{yu2023successor} combined SFs with \emph{predecessor} representations, which encode retrospective information about the agent's trajectory. This ideas, motivated by work in neuroscience \citep{namboodiri2021learning}, was shown to more efficiently target exploration towards bottleneck states (i.e., access points between large regions of the state space).

\subsection{Transfer}\label{sec:ai-transfer}

We've already introduced the idea of cross-task transfer in our discussion of GPI. We now review the broader range of ways in which the challenges of transfer have been addressed using predictive representations.

\subsubsection{Transferring policies between tasks}

We first consider transfer across tasks that are defined by different reward functions. In the following two sections, we consider other forms of transfer.

\paragraph{Few-shot transfer between pairs of tasks.} SFs can enable transferring policies from one reward function $R$ to another function $R_{\tt new}$ by exploiting~\Eqref{eq:sf-q-transfer} with learned cumulant $\cumulant_{\theta}$ and SFs $\sf_{\theta}$ for a source task. At transfer time, one freezes each set of parameters and solves for $\task_{\tt new}$ (e.g., with~\Eqref{eq:reward-loss}).
~\citet{kulkarni2016deep} showed that this enabled an RL agent that learned to play an Atari game to transfer to a new version of that game where the reward function was scaled. Later,~\citet{zhu2017visual} showed that this enabled transfer to new ``go to'' tasks in a photorealistic 3D household environment.

\paragraph{Continual learning across a set of tasks.} Beyond transferring across task pairs, an agent may want to continually transfer its knowledge across a set of tasks.~\citet{barreto2017successor} showed that SFs and GPI provide a natural framework to do this. Consider sequentially learning $n$ tasks.
As the agent learns new tasks, they maintain a growing library of SFs $\{\sf^{\pi_i}\}^{m}_{i=1}$ where $m<n$ is the number of tasks learned so far. When the agent is learning the $m$-th task, they can select actions with GPI according to~\Eqref{eq:sf-gpi} using $\task_m$ as the current transfer task. The agent learns SFs for the current task according to~\Eqref{eq:sf-scalar-loss}.
\citet{zhang2017deep} extended this approach to enable continual learning when the environment state space and dynamics were changing across tasks but still relatively similar. Transferring SFs to an environment with a different state space requires leveraging new state-features. Their solution involved reusing old state-features by mapping them to the new state space with a linear projection. By exploiting linearity in the Q-value decomposition (see~\Eqref{eq:sf-q-transfer}), this allowed reusing SFs for new environments.

\paragraph{Zero-shot transfer to task combinations.} Another benefit of SFs and GPI is that they facilitate transfer to task conjunctions when they are defined as weighted sums of training tasks, $\task_{\tt new} = \sum^n_{i=1} \alpha_i \task_i$. A clear example is combining ``go to'' tasks~\citep{barreto2018transfer,borsa2018universal,barreto2020fast,carvalho2023composing}. For example, consider $4$ tasks $\{\task_1, \ldots, \task_4\}$ defining by collecting different object types; $\task_{\tt new}=(-1)\cdot\task_1 + 2\cdot\task_2 + 1\cdot \task_3 + 0\cdot \task_4$ defines a new task that tries to avoid collecting objects of type $1$, while trying to collect objects of type $2$ twice as much as objects of type $3$.
This approach has been extended to combining policies with continuous state and action spaces~\citep{hunt2019composing}, though it has so far been limited to combining only two policies.
Another important limitation of this general approach is that it can only specify which tasks to prioritize, but cannot specify an \textit{ordering} for these tasks. For example, there is no way to specify collecting object type $1$ \textit{before} object type $2$. One can address this limitation by learning a state-dependent transfer task encoding as with the Option Keyboard (\S\ref{sec:ok}).

\subsubsection{Learning about non-task goals}\label{sec:non-task}

In the previous section, we discussed the transfer setting where an agent learns about a task $\task$ and then subsequently wants to transfer this knowledge to another task $\task_{\tt new}$. In this section, we consider an agent that is learning task $\task$ and wants to \textit{concurrently} learn policies for other tasks (defined by $\tilde{\task}$). That is, each experience trying to accomplish $\task$ is reused to learn how to accomplish $\tilde{\task}$.
This can broadly be categorized as \emph{off-task learning}.

~\citet{borsa2018universal} showed that one can reuse experiences accomplishing task $\task$ to learn control policies for tasks $\tilde{\task}$ that are not too far from $\task$ in the task encoding space by leverage universal SFs. In particular, nearby off-task goals can be sampled from a Gaussian ball around $\task$ with standard deviation $\sigma$: $Z=\{\tilde{\task}_i \sim \mathcal{N}(\task_{\tt new}, \sigma)\}^n_{i=1}$. Then an SF loss following~\Eqref{eq:sf-scalar-loss} would be applied for each non-task goal $\tilde{\task}_i$. Key to this is that the optimal action for each $\tilde{\task}_i$ would be the action that maximized the features determined by $\tilde{\task}_i$ at the next time-step:  $a^* = \argmax_a' \sf_{\theta}(s',a',\tilde{\task}_{i})^{\top}\tilde{\task}_{i}$. This enabled an agent to concurrently learn a policy for not only $\task$ but also for non-task goals $\tilde{\task}_i$ \textit{with no direct experience on those tasks} in a simple 3D navigation environment.

Another example where off-task learning is useful is in \emph{hindsight experience replay}. Typically, experiences that don't accomplish a task $\task$ don't contribute to learning unless some form of reward shaping is employed. Hindsight experience replay provides a strategy for automating reward shaping. In this setting, when the agent fails to accomplish $\task$, it relabels one of the states in its experience as a \textit{fictitious} goal $\tilde{\task}$ for that experience ~\citep{andrychowicz2017hindsight}. This strategy particularly effective when tasks have sparse rewards as it leads there to be a dense reward signal.
When learning a policy with SMs (\S\ref{sec:successor-models}), hindsight experience replay naturally arises as part of the learning objective. It has been shown to improve sample efficiency in sparse-reward virtual robotic manipulation domains and long-horizon navigation tasks~\citep{eysenbach2020c,eysenbach2022contrastive,zheng2023contrastive}. Despite their potential, learning and exploiting SMs is still in its infancy, whereas SFs have been more thoroughly studied. Recently,~\citet{schramm2023usher} developed an asymptotically unbiased importance sampling algorithm that leverages SFs to remove bias when estimating value functions with hindsight experience replay. This enabled learning for both simulation and real-world robotic manipulation tasks in environments with large state and action spaces.

\subsubsection{Other advances in transfer}\label{sec:transfer-advances}

\paragraph{Generalization to new environment dynamics.} One limitation of SFs is that they're tied to the environment dynamics $\envprobs$ with which they were learned.~\citet{lehnert2020successor} and~\citet{han2021option} both attempt to address this limitation by learning SFs over state abstractions which respect bisimulation relations~\citep{li2006towards}.~\citet{abdolshah2021new} attempt to address this by integrating SFs with Gaussian processes such that they can be quickly adapted to new dynamics given a small amount of experience in the new environment. 

\paragraph{Synthesizing new predictions from sets of SFs.} While GPI enables combining a set of SFs to produce a novel policy, it does not generate a novel \textit{prediction} of what features will be experienced from a combination of policies. Some methods attempt to address this by convex combination of SFs~\citep{brantley2021successor,alegre2022optimistic}.

\paragraph{Alternatives to generalized policy improvement.}
~\citet{madarasz2019better} develop the Gaussian SF, which learns a set of reward maps for different environments that can be adaptively combined to adjudicate between different policies. While this compared favorably to GPI, these results were in toy domains; it is currently unclear if their method scales to more complex settings as gracefully as GPI. A potentially more promising alternative to GPI is Geometric Policy Composition ~\citep[GPC;][]{thakoor2022generalised} which enables estimating Q-values when one follows a ordered sequence of $n$ policies $(\pi_1, \ldots, \pi_n)$. Whereas GPI evaluates how much reward will be obtained by the best of a set of policies, GPC is a form of \textit{model-based control} where the agent evaluates the path obtained from following a sequence of policies. We discuss this in more detail in \S\ref{sec:mbrl}.

\subsection{Hierarchical reinforcement learning}\label{sec:hrl}

Many tasks have multiple time-scales, which can be exploited using a hierarchical architecture in which the agent learns and acts at multiple levels of temporal abstraction. The classic example of this is the \emph{options} framework \citep{sutton1999between}. An option $o$ is a temporally-extended behavior defined by (a) an \textit{initiation} function $\mathcal{I}_o$ that determines when it can be activated, (b) a policy $\pi_o$, and (c) a \textit{termination} function $\beta_o$ that determines when the option should terminate: $o = \langle \mathcal{I}_o, \pi_o, \beta_o \rangle$. In this section, we discuss how predictive representations can be used to discover useful options and transfer them across tasks.

\subsubsection{Discovering options to efficiently explore the environment}

One key property of the SR is that it captures the long-range structure of an agent's paths through the environment.
This has been useful in discovering options.
\citet{machado2017eigenoption,machado2023temporal} showed that if one performs an eigendecomposition on a learned SR, the eigenvectors corresponding to the highest eigenvalues could be used to identify states with high diffusion (i.e., states that tend to be visited frequently under a random walk policy). In particular, for eigenvector $e_{i}$, they defined the following intrinsic reward function:
\begin{equation}\label{eq:covering-option-reward}
  R_{i}^{\tt int}(s,s') = e_{i}^{\top} (\cumulant(s') - \cumulant(s)),
\end{equation}
which rewards the agent for exploring diverse parts of the state space. Note that $\cumulant$ is either a one-hot vector in the case of the SR or state-features in the case of SFs. Here, we focus on the SR.
With this reward function, the agent can iteratively build up a set of options $\mathcal{O}$ that can be leveraged to improve exploration of the environment.
The algorithm proceeds as follows:
\begin{enumerate}
  \item \textbf{Collect samples} with a random policy which selects between between primitive actions and options. The set of options is initially empty ($\mathcal{O}=\{\emptyset\}$).
  \item \textbf{Learn successor representation $\sr^{\pi}$} from the gathered samples.
  \item \textbf{Get new exploration option} $o = \langle \mathcal{I}_o, \pi_o, \beta_o \rangle$. $\pi_o$ is a policy that maximizes the intrinsic reward function in~\Eqref{eq:covering-option-reward} using the \textit{current} SR. The initiation function $\mathcal{I}_o$ is $1$ for all states. The termination function $\beta_o$ is $1$ when the intrinsic reward becomes negative (i.e.,~when the agent begins to go towards more frequent states). This option is added to the overall set of options, $\mathcal{O} \leftarrow \mathcal{O} \cup \{o\}$.
\end{enumerate}
 Agents endowed with this strategy were able to discover meaningful options and improve sample efficiency in the four-rooms domain, as well as on challenging Atari games such as Montezuma's revenge.

\subsubsection{Transferring options with the SR}\label{sec:ok-transfer}

\paragraph{Instant synthesis of combinations of options.} One of the benefits of leveraging SFs is that they enable transfer to tasks that are linear combinations of known tasks (see \S\ref{sec:sf-gpi}). At transfer time, an agent can exploit this when transferring options by defining subgoals using this space of tasks (see \S\ref{sec:ok}). In continuous control settings where an agent has learned to move in a set of directions (e.g., up, down, left, right), this has enabled generalization to combinations of these policies~\citep{barreto2019option}. For example, the agent could instantly move in novel directions (e.g., up-right, down-left, etc.) as needed to complete a task.

\paragraph{Transferring options to new dynamics.} One limitation of both options and SFs is that they are defined for a particular environment and don't naturally transfer to new environments with different transition functions.~\citet{han2021option} addressed this limitation by learning \textit{abstract options} with SFs defined over abstract state features that respect bisimulation relations~\citep{li2006towards}. This method assumes transfer from a set of $N$ $\sf$-MDPs $\{\mathcal{M}_i\}_{i=1}^n$ where $\mathcal{M}_i=\langle S_i, O_i, T_i, \sf, \gamma \rangle$. It also assumes the availability of MDP-dependent state-feature functions $\cumulant_{\mathcal{M}_i}(s,a)$ that map individual MDP state-action pairs to a shared feature space. Assume an agent has learned a (possibly) distinct set of options $\{o_k\}$ for each source MDP. The SFs for each option's policy are defined as usual (\Eqref{eq:sf-def}). When we are transferring to a new MDP $\mathcal{M}'$, we can map each option to this environment by mapping its SFs to a policy which produces similar features using an inverse reinforcement learning algorithm~\citep{ng2000algorithms}. Once the agent has transferred options to a new environment, it can plan using these options by constructing an abstract MDP over which to plan.~\citet{han2021option} implement this using \textit{successor homomorphisms}, which define criteria for aggregating states to form an abstract MDP. In particular, pairs of states $(s_1, s_2)$ and options $(o_1, o_2)$ will map to the same abstract MDP if they follow bisimulation relations~\citep{li2006towards} and if their SFs are similar:
\begin{equation}
  || \sf^{o_1}{(s_1, a)} - \sf^{o_2}{(s_2, a)} || \leq \epsilon_{\psi},
\end{equation}
where $\epsilon_{\psi}$ is a similarity threshold. With this method, agents were able to discover state abstractions that partitioned large grid-worlds into intuitive segments and successfully plan in novel MDPs with a small number of learning updates.

\subsection{Jumpy model-based reinforcement learning}\label{sec:mbrl}

The successor model (SM) is interesting because it offers a novel way to do model-based RL. Traditionally, a model-based agent simulates trajectories with a single-step model. While this is flexible, it is also expensive. SMs enable an alternative strategy, where the agent instead samples and evaluates likely (potentially distal) states that will be encountered when following some policy $\pi$. As mentioned in \S\ref{sec:transfer-advances}, \citet{thakoor2022generalised} leverage this property to develop Generalized Policy Composition (GPC), a novel algorithm that enables a jumpy form of model-based RL. Rather than simulate trajectories defined over next states, agents simulate trajectories by using SMs to jump between states using a given set of policies. While this is not as flexible as simulating trajectories with a single-step model, it is much more efficient.

In RL, one typically uses a large discount factor ($\gamma \approx 1$). When learning an SM, this is useful because you can learn likelihoods over potentially very distal states. However, this makes learning an SM more challenging.
GPC mitigates this challenge by composing a shorter horizon SM $\horizon_\beta^\pi$ with a longer horizon SM $\horizon_\gamma^\pi$, where $\beta < \gamma$. Composing two separate SMs with different horizons has the following benefits. An SM with a shorter horizon $\horizon_\beta^\pi$ is easier to learn but cannot sample futures as distal as $\horizon_\gamma^\pi$; on the other hand, $\horizon_\gamma^\pi$ is harder to learn but can make very long-horizon predictions and better avoids compounding errors. By combining the two, \citet{thakoor2022generalised} studied how these two errors can be traded off. 

Intuitively, GPC works as follows.
Given a starting state-action pair $(s_0, a_0)$ and policies $(\pi_1, \ldots, \pi_n)$, the agent samples a sequence of $n-1$ next states with our shorter horizon SM $\horizon_\beta^{\pi_i}$ and $\pi_i$, i.e. $s_1 \sim \horizon_{\beta}^{\pi_1}(\cdot | s_0, a_0)$, $a_1 \sim \pi_1(\cdot|s_1)$, $s_2 \sim \horizon_{\beta}^{\pi_1}(\cdot | s_1, a_1)$, and so on. The agent then samples a (potentially more distal) state $s_n$ from the longer-horizon SM, $\horizon_\gamma^\pi$, $s_n \sim \horizon_{\beta}^{\pi_{n}}(\cdot | s_{n-1}, a_{n-1})$. The reward estimates for the sampled state-action pairs can be combined as a weighted sum to compute $Q^{\pi}(s_0,a_0)$ analogously to~\eqref{eq:successor-policy-eval} \citep[see][for technical details]{thakoor2022generalised}. Leveraging GPC enabled convergence with an order of magnitude fewer samples in the four-rooms domain and in a continuous-control maze navigation domain.

\subsection{Multi-agent reinforcement learning}\label{sec:mulit-agent}

As we've noted earlier, the SR is conditioned on a policy $\pi$. In a single-agent setting, the SR provides predictions about what that agent can expect to experience when executing the policy. In multi-agent settings, one can parametrize this prediction with \textit{another agent's} policy to form predictions about what one can expect to see in the environment when other agents follow their own policies. This is the basis for numerous algorithms that aim to learn about, from, and with other agents~\citep{rabinowitz2018machine,kim2022disentangling,filos2021psiphi,gupta2021uneven,lee2019truly}.

\paragraph{Learning about other agents.}~\citet{rabinowitz2018machine} showed that an AI agent could learn aspects of theory of mind (including passing a false belief test) by meta-learning SFs that described other agents. 
While Rabinowitz did not explicitly compare against humans (and was not trying to directly model human cognition), this remains exciting as a direction for exploring scalable algorithms for human-like theory of mind.

\paragraph{Learning from other agents.} One nice property of SFs is that they can be learned with TD-learning using \textit{off-policy} data (i.e.,~data collected from a policy different from the one currently being executed). 
This can be leveraged to learn SFs for the policies of other agents just as an agent learns SFs for their own policy.~\citet{filos2021psiphi} exploited this to design an agent that simultaneously learned SFs for both its own policy and for multiple other agents. They were then able to generalize effectively to new tasks via a combination of all of their policies by exploiting GPI (\Eqref{eq:sf-gpi}).

\paragraph{Learning with other agents.} One of the key benefits of leveraging universal SFs with GPI (\Eqref{eq:usfa-gpi}) is that you can systematically change the Q-value used for action selection by (1) shifting the policy encoding $\behavior$ defining the SF, or (2) shifting the feature preferences $\task$ that are being maximized.~\citet{gupta2021uneven} exploit this for cooperative multi-agent RL. Let $s^{(i)}$ denote the state of the $i$-th agent and $a^{(i)} \in \mathcal{A}_i$ denote the action they take.
Now let $\sf_{\theta}\left(s^{(i)}, a, \behavior\right)$ denote that agent's SFs for policy encoding $\behavior$, let $\mathcal{C}_{\tt tasks}$ denote the set of possible feature preferences that agents can follow, and let $\mathcal{C}_{\pi}$ denote the set of possible policy encodings over which agents can make predictions. \citet{gupta2021uneven} studied the cooperative setting where the overall Q-value is simply the sum of individual Q-values.
If $\mathcal{C}_{\pi}$ denotes policies for separate but related tasks, acting in accordance to
\begin{equation}\label{eq:sf-gpi-multiagent}
  a^{(i)} = \argmax_{a \in \mathcal{A}_i} \max_{\task \in \mathcal{C}_{\tt tasks}} \max_{\behavior \in \mathcal{C}_{\pi}}\left\{\sf_{\theta}\left(s^{(i)}, a, \behavior\right)^{\top} \task\right\}
\end{equation}
enables a team of agents to take actions which systematically explore the environment on tasks specified by $\mathcal{C}_{\tt tasks}$.
These teams were able to improve exploration and zero-shot generalization in a variety of cooperative multi-agent environments including Starcraft~\citep{samvelyan2019starcraft}.

\subsection{Other artificial intelligence applications}

The SR and its generalizations have been broadly applied within other areas of AI. For example, it has been used to define an improved similarity metric in episodic control settings~\citep{emukpere2021successor}. By leveraging SFs, one can incorporate information from previously experienced states with similar dynamics to the current state. The SR has also been applied towards improving importance sampling in off-policy learning. If the agent learns a density ratio similar to~\Eqref{eq:successor-ratio}, this can enable simpler marginalized importance sampling algorithms~\citep{liu2018breaking} that improve off-policy evaluation~\citep{fujimoto2021deep}. In addition to these examples, we highlight the following applications of the SR.

\paragraph{Representation learning.} Learning SMs can obviate the need for separate representation losses.
In many applications, the reward signal is not enough to drive learning of useful representations. Some strategies to address this challenge include data augmentation and learning of auxiliary tasks. Learning the SM has been shown to enable representation learning with superior sample efficiency without using these additions~\citep{eysenbach2020c,eysenbach2022contrastive,zheng2023contrastive}. Predictive representations can also be used to define an auxiliary task for representation learning, which has been shown to  be helpful in several settings. A simple example comes from inspecting the loss for learning SFs (\Eqref{eq:sf-scalar-loss}). In standard Q-learning, the agent only learns about achieving task-specific reward. When learning SFs, the agent also learns representations that enable achieving state-features that are potentially not relevant for the current task (i.e., the agents are by default leaning auxiliary tasks). This important ability is even possible in a continual learning setting where the distribution of state-features is non-stationary~\citep{mcleod2021continual}. Another interesting example comes from proto-value networks~\citep{farebrother2023proto}. The authors show that if one learns a successor measure (a set-inclusion based generalization of the SR) over random sets, this can enable the discovery of predictive representations that enable very fast learning in the Atari learning environment.

\paragraph{Learning diverse policies.} A final application of the SR has been in learning diverse policies. In the mujoco environment,~\citet{zahavy2021discovering} showed that SFs enabled discovering a set of diverse policies for controlling a simulated dog avatar. Their approach used SFs to prospectively summarize trajectories. A set of policies was than incrementally learned so that each new policy would be different in its expected features from all policies learned so far. This ideas was then generalized to diversify chess playing strategies based on their expected future features~\citep{zahavy2023diversifying}. %

\section{Neuroscience applications}

In this section, we discuss how the computational ideas reviewed above have been used to understand a variety of brain systems. Medial temporal lobe regions, and in particular the hippocampus, appear to encode predictive representations. In the following subsections, we review evidence for this claim and efforts to formalize its mechanistic basis in neurobiology. We also discuss how vector-valued dopamine signals may provide an appropriate learning signal for these representations.

\subsection{A brief introduction to the medial temporal lobe}

Before discussing evidence for predictive representations, it is important to take a sufficiently broad view of the medial temporal lobe's functional organization. Not everything we know about these regions fits neatly into a theory of predictive representations. Indeed, classical views are quite different, emphasizing spatial representation and episodic memory.

\begin{figure}
\centering
\includegraphics[width=\textwidth]{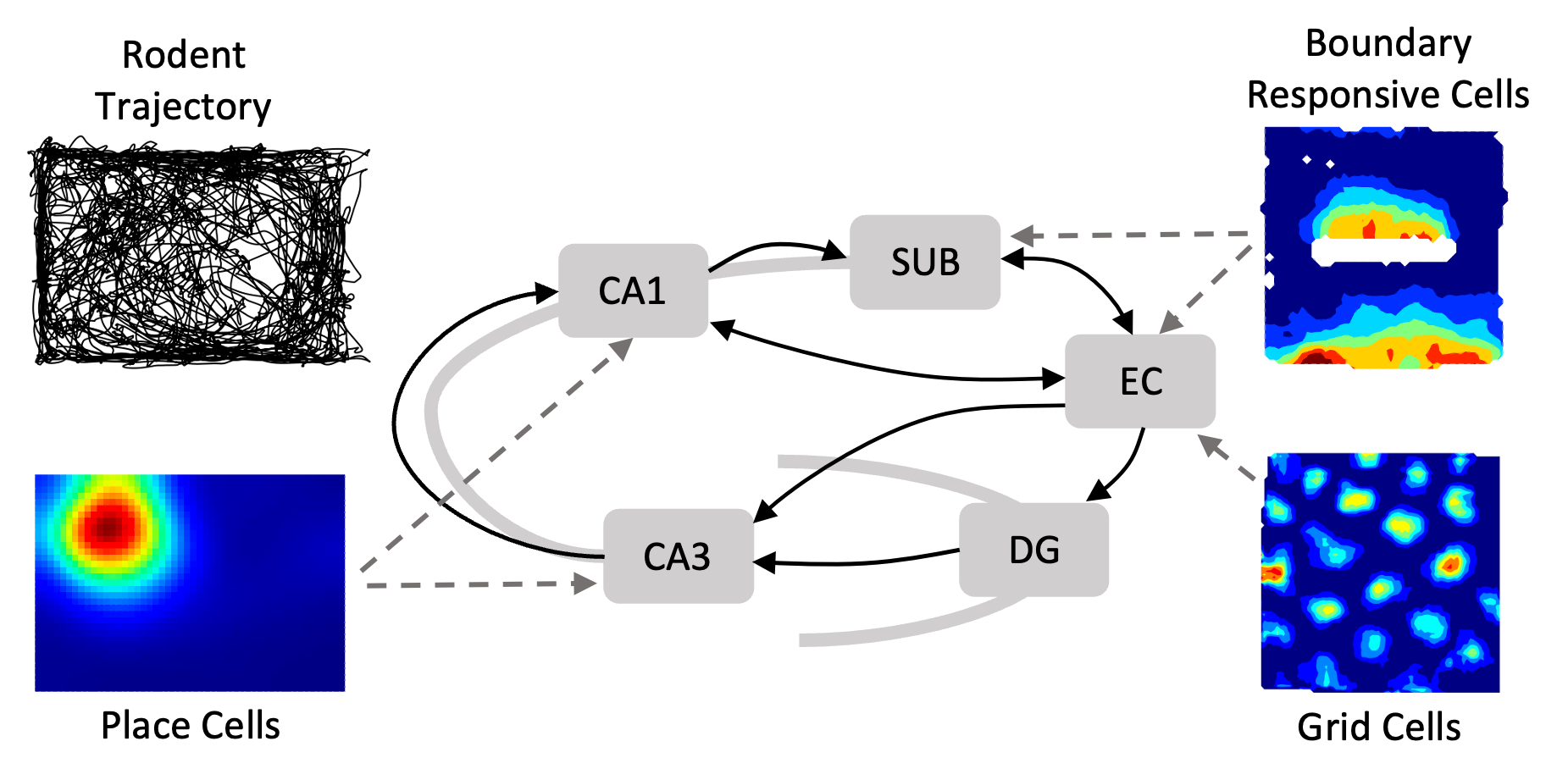}
\caption{\textbf{Spatial representations in the medial temporal lobe.} As a rodent navigates space (e.g., a rectangular arena; top left), place cells recorded in regions CA1 and CA3 of Hippocampus fire in stable, sparse representations of self-location (bottom left; hot colors indicate increased neuronal activity). Conversely, grid cells in neighbouring medial entorhinal cortex (EC; bottom right), which provides input to both CA1 and CA3 directly,  as well as CA3 via dentate gyrus (DG; solid arrows imply directional connectivity between regions), have spatially periodic hexagonal firing patterns tiling the entire environment. Additionally, boundary responsive neurons in both medial entorhinal cortex and subiculum (SUB) fire when the animal occupies specific positions relative to external and internal environmental boundaries.
}
\label{fig:neuroanatomy}
\end{figure}

Extensive evidence identifies the hippocampus and associated cortical regions as providing a neural-level representation of space, often conceptualized as a cognitive map \citep[][see also \S\ref{sec:spatial}]{okeefeHippocampusCognitiveMap1978,morris1982place}. Integral to this framework are the distinct firing patterns of various cell types found in structures across the hippocampal formation (Figure \ref{fig:neuroanatomy}). Place cells, in regions CA3 and CA1, offer a temporally stable, sparse representation of self-location able to rapidly reorganize in novel environments---the phenomenon of remapping \citep{o1971hippocampus,muller1987effects,bostock1991experience}.

Subiculum and dentate gyrus also contain spatially modulated neurons with broadly similar characteristics, the former tending to be diffuse and elongated along environmental boundaries \citep{lever2009boundary} while the latter are extremely sparse \citep{jung1993spatial,leutgeb2007pattern}. In contrast, in medial entorhinal cortex (mEC), the primary cortical partner of hippocampus, the spatially periodic firing patterns of grid cells effectively tile the entire environment  (Figure \ref{fig:neuroanatomy}) and are organized into discrete functional modules of different scale \citep{hafting2005microstructure,barry2007experience,stensola2012entorhinal}. The highly structured activity of grid cells has provoked a range of theoretical propositions pointing to roles in path integration \citep{mcnaughton2006path,burgess2007oscillatory}, vector based navigation \citep{bush2015using,banino2018vector}, and as an efficient basis set for spatial generalization \citep{whittington2020tolman}. Notably, mEC also contains a `zoo' of other cell types with functional characteristics related to self-location, including head direction cells \citep{sargolini2006conjunctive}, border cells \citep{solstad2008representation}, speed cells \citep{kropff2015speed}, and multiplexed conjunctive responses \citep{sargolini2006conjunctive,hardcastle2017multiplexed}.

In summary, the medial temporal lobe exhibits remarkable functional diversity. We now turn to the claim that predictive principles offer a unifying framework for understanding aspects of this diversity.

\subsection{The hippocampus as a predictive map}  \label{sec:hippocampus}

\begin{figure}
\centering
\includegraphics[width=\textwidth]{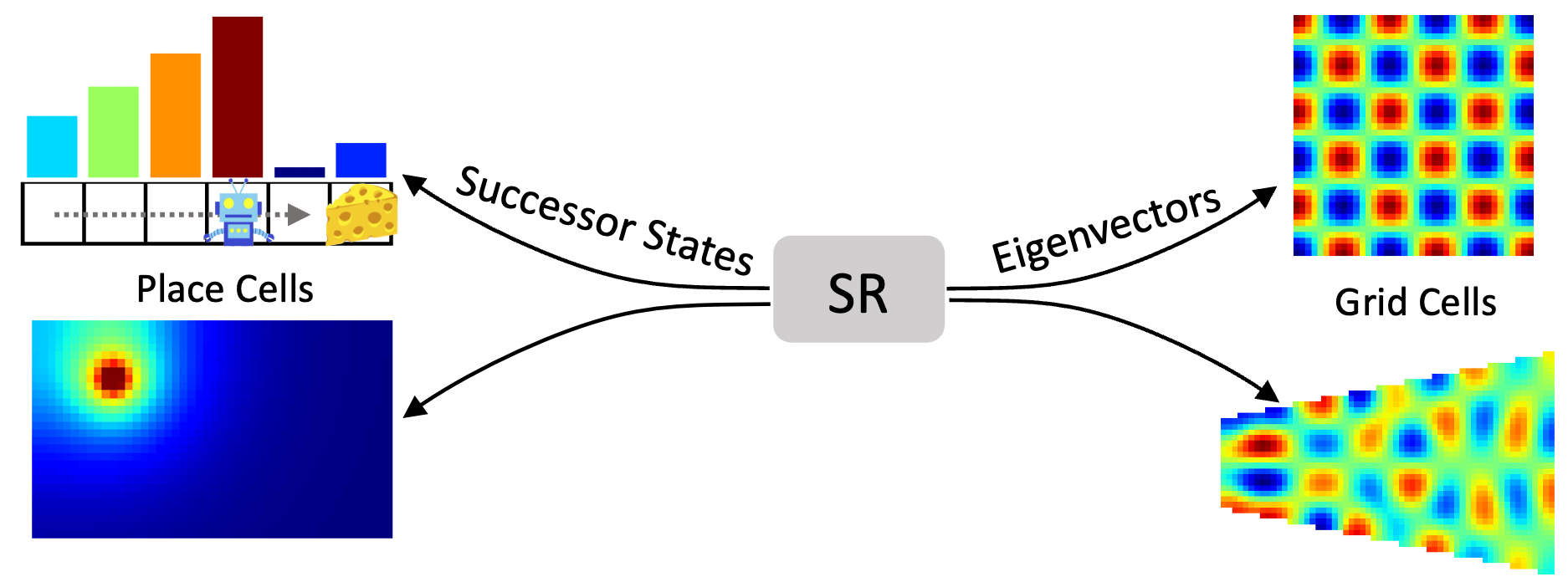}
\caption{\textbf{Successor representation model of the hippocampus and medial entorhinal cortex.} As an agent explores a linear track environment in a uni-directional manner, the SR skews backwards down the track opposite to the direction of motion (top left; hot colors indicate increased predicted occupancy of the agent's depicted state), as observed in hippocampal place cells \citep{mehta2000experience}. In a 2D arena, the SR forms place cell-like sparse representations of self-location (bottom left), while the eigenvectors of the SR form spatially periodic activity patterns reminiscent of entorhinal grid cells (top right). Similarly to real grid cells \citep{krupic2015grid}, the periodicity of these eigenvectors is deformed in polarized environments such as trapezoids (bottom right).
}
\label{fig:neuro_sr_gridworld}
\end{figure}

Accumulating evidence indicates that neurons within the hippocampus and its surrounding structures, particularly place and grid cells, demonstrate predictive characteristics consistent with a predictive map of spatial states. \cite{stachenfeld2017hippocampus} were the first to systematically explore this perspective, establishing a connection between the responses of hippocampal neurons and the SR.\footnote{Earlier ideas about predictive processing in the hippocampus were explored by several authors \citep{levy1996sequence,levy2005interpreting,buckner2010role,lisman2009prediction}, though these were not framed in terms of the SR or related ideas.} They argued that place cells were not inherently representing the animal's spatial location, but rather its expectations about future spatial locations. Specifically, they argued that the receptive fields of places cells correspond to columns of the SR matrix $\mathbf{M}^\pi$ from \S\ref{sec:sr} (Figure \ref{fig:neuro_sr_gridworld}, left). This implies that each receptive field is a \emph{retrodictive code}, in the sense that the cells are more active in locations that tend to precede the cell's ``preferred'' location (i.e., the location of the peak firing). The population activity of place cells at a given time corresponds to a row of the SR matrix; this is a \emph{predictive code}, in the sense that they collectively encode expectations about upcoming states.

In line with this hypothesis, the tuning of place fields (Figure \ref{fig:neuro_sr_gridworld}, bottom left) are influenced by the permissible transitions afforded by an environment: they do not typically cross environmental boundaries like walls, tending to extend along them, mirroring the trajectories animals follow \citep{alvernhe2011local,tanni2022state}. Alternations made to an environment's layout, affecting the available paths, influence the activity of adjacent place cells, consistent with the SR \citep{stachenfeld2017hippocampus}. Notably, even changes in policy alone, such as training rats to switch between foraging and directed behaviour, can markedly alter place cell firing \citep{markus1995interactions}, also broadly consistent with the SR. Further, when animals are trained to generate highly stereotyped trajectories, for example repeatedly traversing a track in one direction, CA1 place fields increasingly exhibit a backwards skew, opposite to the direction of travel \citep{mehta2000experience}, thereby anticipating the animal's future state. This arises naturally from learning the SR, since the upcoming spatial states become highly predictable when agents consistently move in one direction, resulting in a backwards skew of the successor states (Figure \ref{fig:neuro_sr_gridworld}, top left). The basic effect is captured in simple grid worlds like those used by \cite{stachenfeld2017hippocampus}, but when the anchoring is replaced with continuous feature-based methods, the successor features also capture the backwards shift in field peak observed in neural data \citep{mehta2000experience, george2023rapid}.  

While the properties of place cells are consistent with encoding the SR, grid cells appear to resemble the eigenvectors of the SR (Figure \ref{fig:neuro_sr_gridworld}, right). Specifically, eigendecomposition of the SR matrix $\mathbf{M}^\pi$ yields spatially periodic structures of varying scales, heavily influenced by environmental geometry \citep{stachenfeld2017hippocampus} while being relatively robust to the underlying policy \citep{de2020neurobiological}. Broadly, these resemble grid cells, but notably lack the characteristic hexagonal periodicity, except when applied to hexagonal environments. This discrepancy, however, is likely not significant because subsequent work indicates that biological constraints, such as non-negative firing rates \citep{dordek2016extracting,sorscher2019unified}, efficiency considerations \citep{dorrell2023actionable}, and neurobiologically plausible state-features \citep{de2020neurobiological} tend to move these solutions closer to the expected hexagonal activity patterns (Figure \ref{fig:neuro_bvc_sr}A). The key point then is that environmental geometries that polarize the transitions available to an animal produce SR eigenvectors with commensurate distortions, matching observations that grid firing patterns are also deformed under such conditions \citep{derdikman2009fragmentation,krupic2015grid}. Notably, this phenomenon is also observed in open-field environments with straight boundaries, where biological grid cells and SR-derived eigenvectors both exhibit a tendency to orient relative to walls \citep{krupic2015grid,de2020neurobiological}. Complementary evidence comes from virtual reality studies of human subjects, where errors in distance estimates made by participants mirrored distortions in eigenvector-derived grid cells \citep{bellmund2020deforming}.

Although place and grid cells are predominantly conceptualized as spatial representations, it is increasingly clear that these neurons also represent non-spatial state spaces \citep{constantinescu2016organizing,aronov2017mapping}; in some cases, activity can be interpreted as encoding an SR over such state spaces. For example, a study by \cite{garvert2017map} showed human participants a series of objects on a screen in what appeared to be a random order. However, unknown to the participants, the sequence was derived from a network of non-spatial relationships, where each object followed certain others in a predefined pattern. Brain imaging found that hippocampal and entorhinal activity mirrored the non-spatial predictive relationships of the objects, as if encoded by an SR \citep[see also][]{brunec2022predictive}.

\subsection{Learning a biologically plausible SR}\label{sec:bio-sr}

In much of the neuroscience work, SRs are formulated over discrete state spaces, facilitating analysis and enabling direct calculation for diffusive trajectories. In spatial contexts, this corresponds to a grid world with one-hot location encoding, a method that can produce neurobiologically plausible representations \citep{stachenfeld2017hippocampus}. However, the brain must use biologically plausible learning rules and features derived from sensory information, with the choice of state-features $\cumulant(s)$ exerting significant influence on the resultant SFs $\sf^\pi(s)$ (described in \S\ref{sec:sf}).

\begin{figure}
\centering
\includegraphics[width=\textwidth]{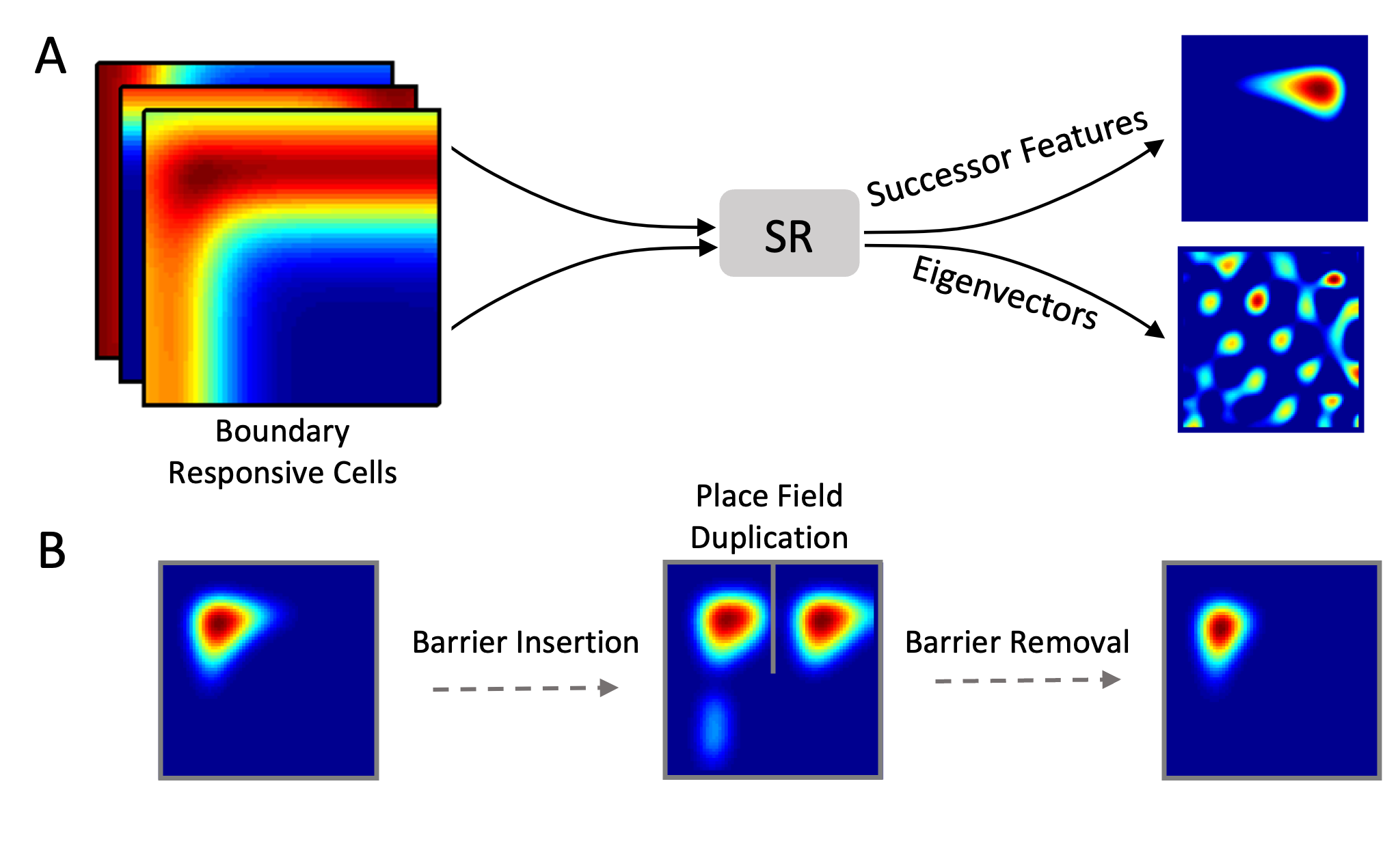}
\caption{\textbf{A successor representation with neurobiological state features.} (A) Boundary responsive cells, present in subiculum and entorhinal cortex, are used as state features for learning successor features and their eigenvectors \citep{de2020neurobiological}. These can capture more nuanced characteristics of both place and grid cells compared to one-hot spatial features derived from a grid world. For example, the eigenvectors have increased hexagonal periodicity relative to a grid world SR model. (B) The successor features can convey duplicated fields when additional walls are inserted in the environment, as observed in real place cells \citep{barry2006boundary}. 
}
\label{fig:neuro_bvc_sr}
\end{figure}

\cite{de2020neurobiological} employed idealized boundary vector cells (BVCs)---neurons coding for distance and direction to enviromental boundaries---as a basis over which to calculate a spatial SR. BVCs have been hypothesized as inputs to place cells \citep{hartley2000modeling}. They resemble the boundary-responsive cells found in the mEC \citep{solstad2008representation} and subiculum \citep{barry2006boundary,lever2009boundary}; hence, they are plausibly available to hippocampal circuits (Figure \ref{fig:neuro_bvc_sr}A). The SFs of these neurobiological state-features and their eigendecomposition resemble place and grid fields, as before, but also captured more of the nuanced characteristics of these spatially tuned neurons---for example, the way in which place fields elongate along  environmental boundaries \citep{tanni2022state} and duplicate when additional walls are introduced \citep[Figure \ref{fig:neuro_bvc_sr}B;][]{barry2006boundary}. \cite{geerts2020general} employed a complementary approach, using an SR over place cell state-features in parallel with a model-free framework trained on egocentric features. The dynamics of the interaction between these two elements mirrored the behavioral preference of rodents, which initially favor a map-based navigational strategy, before switching to one based on body turns \citep{geerts2020general,packard1996inactivation}.

Subsequent models expanded on these ideas. While these differ in terms of implementation and focus, they employ the common idea of embedding transition probabilities into the weights of a network using biologically plausible learning rules. \cite{fang2023neural} advanced a framework using a bespoke learning rule acting on a recurrent neural network supplied with features derived from experimental recordings. The network was sufficient to calculate an SR and could do so at different temporal discounts, producing SFs that resembled place cells. \cite{bono2023learning} applied spike-time dependent plasticity \citep[STDP;][]{bi1998synaptic,kempter1999hebbian}, a Hebbian learning rule sensitive to the precise ordering of pre- and post-synaptic spikes, to a single-layer spiking network. Because the ordering of spikes from spatial inputs inherently reflects the sequence of transitions between them, this configuration also learns an SR. Indeed, the authors were able to show that the synaptic weights learnt by this algorithm are mathematically equivalent to TD learning with an eligibility trace. Furthermore, temporally accelerated biological sequences, such as replay \citep{wilson1994reactivation,olafsdottir2016coordinated}, provide a means to quickly acquire SRs for important or novel routes. Finally, \cite{george2023rapid} followed a similar approach, showing that STDP \citep{bi1998synaptic} applied to theta sweeps---the highly ordered sequences of place cell spiking observed within hippocampal theta cycles \citep{o1993phase,foster2007hippocampal}---was sufficient to rapidly learn place field SFs that were strongly modulated by agent behavior, consistent with empirical observations. Additionally, because the speed and range of theta sweeps is directly linked to the size of the underlying place fields, the authors also noted that the gradient of place field sizes observed along the dorsal-ventral axis of the hippocampus inherently approximate SFs with decreasing temporal discounts \citep{kjelstrup2008finite,momennejad2018predicting}.

The formulation used by \cite{george2023rapid} highlights a paradox: while place fields can serve as state-features, they are also generated as SFs. This dual role might suggest a functional distinction between areas such as CA3 and CA1, with CA3 potentially providing the spatial basis and CA1 representing SFs. Alternatively, spatial bases could originate from upstream circuits, such as mEC, as proposed in the \cite{fang2023neural} model. Furthermore, it is conceivable that the initial rapid formation of place fields is governed by a distinct plasticity mechanism, such as behavioral-timescale plasticity \citep{bittner2017behavioral}. Once established, these fields would then serve as a basis for subsequent SF learning. Such a perspective is compatible with observations that populations of place fields in novel environments do not immediately generate theta sweeps \citep{silva2015trajectory}.

These algorithms learn SFs under the premise that the spatial state is fully observable, for example by a one-hot encoding in a grid world or the firing of BVCs computed across the distances and directions to nearby walls. However, in reality states are often only partially observable and inherently noisy due to sensory and neural limitations. \cite{vertes2019neurally} present a mechanism for how the SR can be learnt in partially observable noisy environments, where state uncertainty is represented in a feature-based approximation via distributed, distributional codes. This method supports RL in noisy or ambiguous environments, for example navigating a corridor of identical rooms.
    
In summary, the SR framework can be generalized to a state space comprised of continuous, non-identical, overlapping state-features, and acquired with biological learning rules. As such, it is plausible that hippocampal circuits could in principle instantiate a predictive map, or close approximation, based on the mixed population of spatially modulated neurons available from its inputs. The models reviewed above demonstrate ways in which the SR could be learnt online during active experience. Nonetheless, much learning in the brain is achieved offline, during periods of wakeful rest or sleep. One candidate neural mechanism for this is replay, rapid sequential activity during periods of quiescence.

\subsection{Replay}

During periods of sleep and awake rest, hippocampal place cells activate in rapid sequences that recapitulate past experiences \citep{wilson1994reactivation,foster2007hippocampal}. These reactivations, known as replay, often coordinate with activity in the entorhinal \citep{olafsdottir2016coordinated} and sensory cortices \citep{ji2007coordinated,rothschild2017cortical}, and are widely thought to be a core mechanism supporting system-level consolidation of experiential knowledge \citep{girardeau2009selective,ego2010disruption,olafsdottir2015hippocampal}. In linear track environments, hippocampal place cells typically exhibit directional tuning with firing fields that disambiguate travel in either direction \citep{navratilova2012experience}. This directionality enables two functional classes of replay to be distinguished: `forward replay' where the sequence reflects the order in which the animal experienced the world, and `reverse replay' where the behavioural sequence is temporally reversed (analogous to the animal walking tail-first down the track). Intriguingly, the phase of a navigation task has been shown to influence the type of replay that occurs; for example reverse replay is associated with receipt of reward at the end of trials, while forward replay is more abundant at the start of trials prior to active navigation \citep{diba2007forward}.

\cite{mattar2018prioritized} modelled the emergence of forward and reverse replays using a reinforcement learning agent that accesses memories of locations in an order determined by their expected utility. Specifically, the agent prioritizes replaying memories as a balance of two requirements: the `need' to evaluate imminent choices versus the `gain' from propagating newly encountered information to preceding locations. The need term for a spatial state corresponds to its expected future occupancy given the agent's current location, thus utilizing the definition of the SR (\S\ref{sec:sr}, Eq. \ref{eq:sr}) to provide a measure for how often in the near future that state will tend to be visited. The gain term represents the expected increase in reward from visiting that state. This mechanism produces sequences that favor adjacent backups: upon discovery of an unexpected reward, the last action executed will have a positive gain, making it a likely candidate for replay. Thus, value information can be propagated backward by chaining successive backups in the reverse direction, simulating reverse replay. Conversely, at the beginning of a trial, when the gain differences are small and the need term dominates, sequences that propagate value information in the forward direction will be the likely candidates for replay, prioritizing nearby backups that extend forward through the states the agent is expected to visit in the near future.

Sequential neural activity in humans has similarly been observed to exhibit orderings consistent with sampling based on future need. Using a statistical learning task with graph-like dependencies between visual cues \citep{schapiro2013neural,garvert2017map,lynn2020abstract}, \cite{wittkuhn2022statistical} showed participants a series of animal images drawn from a ring-like graph structure in either a uni- or bi-directional manner. Using fMRI data recorded during 10s pauses between trials, \cite{wittkuhn2022statistical} found forward and reverse sequential activity patterns in visual and sensorimotor cortices, a pattern well-captured by an SR model learnt from the graph structure participants experienced.

\subsection{Dopamine and generalized prediction errors}

As described in \S\ref{sec:sr}, TD learning rules provide a powerful algorithm for value estimation. The elegant simplicity of this algorithm led neuroscientists to explore if, and how, TD learning might be implemented in the brain. Indeed, one of the celebrated successes of neuroscience has been the discovery that the activity of midbrain dopamine neurons appears to report reward prediction errors \citep{schultz1997neural} consistent with model-free RL algorithms (\S\ref{sec:algorithmic}, \Eqref{eq:Qlearn}). This successfully accounts for many aspects of dopamine responses in classical and instrumental conditioning tasks \citep{starkweather2021dopamine}.

While elegant, the classical view that dopamine codes for a scalar reward prediction error does not explain more heterogeneous aspects of dopamine responses. For example, the same dopamine neurons also respond to novel and unexpected stimuli \citep[e.g.,][]{ljungberg1992responses,horvitz2000mesolimbocortical}, and to errors in predicting the features of rewarding events, even when value remains unchanged \citep{chang2017optogenetic,takahashi2017dopamine,stalnaker2019dopamine,keiflin2019ventral}. \cite{russek2017predictive} highlighted the biological plausibility of SR TD learning rule (\Eqref{eq:SRTDupdate}-\ref{eq:SRTDerror}) in light of its similarity to the model-free TD learning rule (\Eqref{eq:Qlearn}), while refraining from making any explicit connections between vector-valued SR TD errors and dopamine. \cite{gardner2018rethinking} took this idea further and proposed an extension to the classic view of dopamine, suggesting that it also encodes prediction errors related to sensory features. According to this model, dopamine reports vector-valued TD errors suitable for updating SFs (\S\ref{sec:sf}), utilizing the fact that SFs obey a Bellman equation and hence are learnable by TD. This model explains a number of phenomena that are puzzling under a classic TD model based only on scalar reward predictions.

First, it explains why the firing rate of dopamine neurons increases after a change in reward identity, even when reward magnitude is held fixed \citep{takahashi2017dopamine}: changes in reward identity induce a sensory prediction error that shows up as one component of the error vector. Second, it explains, at least partially, why subpopulations of dopamine neurons encode a range of different non-reward signals \citep{engelhard2019specialized,dejong2022mesoaccumbal,gonzalez2023ventral}. Third, it explains why optogenetic manipulations of dopamine influence conditioned behavior even in the absence of reward \citep{chang2017optogenetic,sharpe2017dopamine}.

How do dopamine neurons encode a vector-valued error signal? One possibility is that the errors are distributed across population activity. Pursuing this hypothesis, \citet{stalnaker2019dopamine} analyzed the information content of dopamine neuron ensembles. They showed that reward identity can be decoded from these ensembles, but not from single neuron activity. Moreover, they showed that this information content disappeared over the course of training following an identity switch, consistent with the idea that error signals go to 0 as learning proceeds. The question remains how a vector-valued learning system is implemented biophysically. Some progress in this direction (albeit within a different theoretical framework), has been made by \citet{warnberg2023feasibility}.

\section{Cognitive science applications}

A rich body of work dating back over a century has linked RL algorithms to reward-based learning processes in humans and nonhuman animals \citep{niv2009reinforcement}. Empirical findings align with the theoretical properties of model-based and model-free control (described in \S\ref{sec:algorithmic}), suggesting that model-based control underlies reflective, goal-directed behaviors, while model-free control underlies reflexive, habitual behaviors. The existence of both systems in the brain and their synergistic operation has received extensive support by a wide range of behavioral and neural studies across a number of species and experimental paradigms \citep[see][for a detailed review]{dolan13}.

Recall that the SR (described in \S\ref{sec:sr}) occupies an intermediate ground between model-based and model-free algorithms. This can make it advantageous when flexibility and efficiency are both desirable, which is the case in most real-world decision making scenarios. In the field of cognitive science, several lines of research suggest that human learning and generalization is indeed consistent with the SR and related predictive representations. In this section, we examine studies showing that patterns of responding to changes in the environment \citep{momennejad2017successor}, transfer of knowledge across tasks \citep{tomov2021multi}, planning in spatial domains \citep{geerts2023probabilistic}, and contextual memory and generalization \citep{gershman2012successor,smith2013context,zhou2023episodic} exhibit signature characteristics of the SR-like predictive representations that cannot be captured by pure model-based or model-free strategies.

\begin{figure}
\centering
\includegraphics[width=\textwidth]{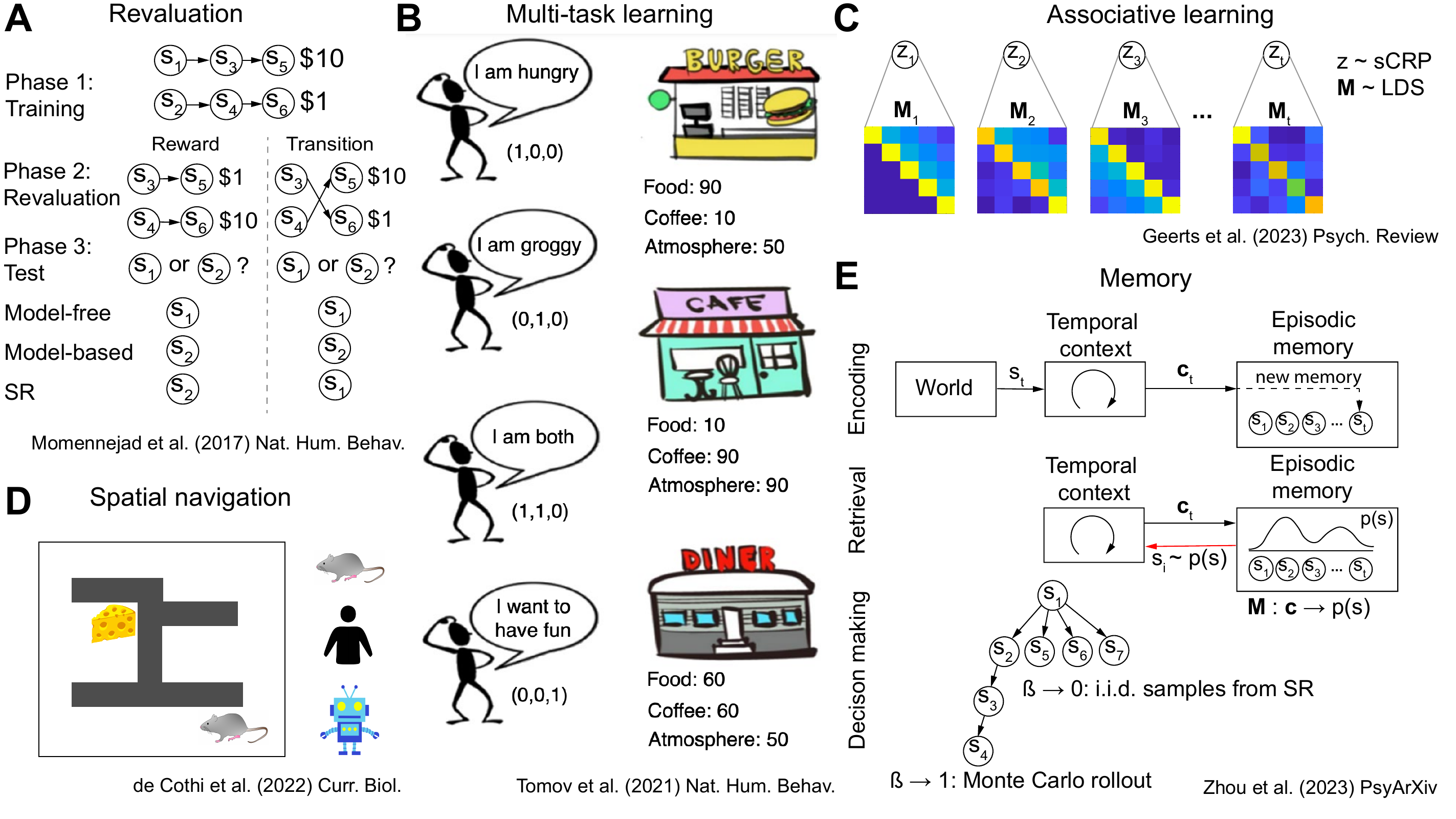}
\caption{\textbf{Predictive representations in cognitive science}. (A) Revaluation paradigm and predictions from \citet[][\S\ref{sec:revaluation}]{momennejad2017successor}. (B) Multi-task learning paradigm from \citet[][\S\ref{sec:MTRL}]{tomov2021multi}. A person is trained on tasks where they are either hungry ($\mathbf{w}_\textit{train}^1 = [1, 0, 0]$) or groggy ($\mathbf{w}_\textit{train}^2 = [0, 1, 0]$), and then tested on a task in which they are hungry, groggy, and looking to have fun ($\mathbf{w}_\textit{new} = [1, 1, 1]$). (C) Context-dependent Bayesian SR from \citet[][\S\ref{sec:associative}]{geerts2023probabilistic}. $z$, context. $\mathbf{M}$, SR matrix associated with each context. sCRP, sticky Chinese restaurant process. LDS, linear-Gaussian dynamical system. (D) Spatial navigation paradigm from \citet[][\S\ref{sec:spatial}]{de2022predictive}. (E) TCM-SR: Using the temporal context model (TCM) to learn the SR for decision making \citep[][\S\ref{sec:memory}]{zhou2023episodic}. $\mathbf{M}$, SR matrix. $\mathbf{c}$, context vector. $s$, state and/or item. SR, successor representation.
}
\label{fig:sr-cogsci}
\end{figure}

\subsection{Revaluation}  \label{sec:revaluation}

Some of the key findings pointing to a balance between a goal-directed system and a habitual system in the brain came from studies of reinforcer revaluation \citep{adams1981instrumental,adams1982variations,dickinson1985actions,holland2004relations}. In a typical revaluation paradigm, an animal (e.g., a rat) is trained to associate a neutral action (e.g., a lever press) with an appetitive outcome (e.g., food). The value of that outcome is subsequently reduced (e.g., the rat is satiated, so food is less desirable) and the experimenter measures whether the animal keeps taking the action in the absence of reinforcement. Goal-directed control predicts that the animal would not take the action, since the outcome is no longer valuable, while habitual control predicts that the animal would keep taking the action, since the action itself was not devalued. Experimenters found that under some conditions---such as moderate training, complex tasks, or disruptions to dopamine inputs to striatum---behavior appears to be goal-directed (e.g., lever pressing is reduced), while under other conditions---such as extensive training, or disruptions to prefrontal cortex---behavior appears to be habitual (e.g., the rat keeps pressing the lever).

A modeling study by \citet{daw2005uncertainty} interpreted these findings through the lens of RL (see \S\ref{sec:algorithmic}). The authors formalized the goal-directed system as a model-based controller putatively implemented in prefrontal cortex and the habitual system as a model-free controller putatively implemented in dorsolateral stratum. They proposed that the brain arbitrates dynamically between the two controllers based on the uncertainty of their value estimates, preferring the more certain (and hence likely more accurate) estimate for a given action. Under this account, moderate training or complex tasks would favor the model-based estimates, since the model-free estimates may take longer to converge and hence be less reliable. On the other hand, extensive training would favor the model-free estimates, since they will likely have converged and hence be more reliable than the noisy model-based simulations. 

One limitation of this account is that it explains sensitivity to outcome revaluation in terms of a predictive \textit{model}, but it does not rule out the possibility that the animals may instead be relying on a predictive \textit{representation}.  A hallmark feature of predictive representations is that they allow an agent to adapt quickly to changes in the environment that keep its cached predictions relevant, but not to changes that require updating them.  In particular, an agent equipped with the SR (\S\ref{sec:sr}) should adapt quickly to changes in the reward structure ($R$) of the environment, but not to changes in the transition structure ($T$). Since the earlier studies on outcome revaluation effectively only manipulated reward structure, both model-based control and the SR could account for them, leaving open the question of whether outcome revaluation effects could be fully explained by the SR instead. 

This question was addressed in a study by \citet{momennejad2017successor} which examined how changes in either the reward structure or the transition structure experienced by human participants affect their subsequent choices. The authors used a two-step task consisting of three phases: a learning phase, a relearning (or revaluation) phase, and a test phase (Figure~\ref{fig:sr-cogsci}A). During the learning phase, participants were presented with two distinct two-step sequences of stimuli and rewards corresponding to two distinct trajectories through state space. The first trajectory terminated with high reward ($s_1 \rightarrow s_3 \rightarrow s_5$ \$10), while the second trajectory terminated with a low reward ($s_2 \rightarrow s_4 \rightarrow s_6$ \$1), leading participants to prefer the first one over the second one. 

During the revaluation phase, participants had to relearn the second half of each trajectory. Importantly, the structure of the trajectories changed differently depending on the experimental condition. In the \textit{reward revaluation} condition, the transitions between states remained unchanged, but the rewards of the two terminal states swapped ($s_3 \rightarrow s_5$ \$1; $s_4 \rightarrow s_6$ \$10). In contrast, in the \textit{transition revaluation} condition, the rewards remained the same, but the transitions to the terminal states swapped ($s_3 \rightarrow s_6$ \$1; $s_4 \rightarrow s_5$ \$10). 

Finally, in the test phase, participants were asked to choose between the two initial states of the two trajectories ($s_1$ and $s_2$). Note that, under both revaluation conditions, participants should now prefer the initial state of the second trajectory ($s_2$) as it now leads to the higher reward. 

Unlike previous revaluation studies, this design clearly disambiguates between the predictions of model-free, model-based, and SR learners. Since the initial states ($s_1$ and $s_2$) never appear during the revaluation phase, a pure model-free learner would not update the cached values associated with those states and would still prefer the initial state of the first trajectory ($s_1$). On the other hand, a pure model-based learner would update its reward ($\hat{R}$) or transition ($\hat{T}$) estimates during the corresponding revaluation phase, allowing it to simulate the new outcomes from each initial state and make the optimal choice ($s_2$) during the test phase. Critically, both model-free and model-based learners \citep[and any hybrid between them, such as a convex combination of their outputs, as in][]{daw2011model} would exhibit the same preferences during the test phase in both revaluation conditions.

In contrast, an SR learner would show differential responding in the test phase depending on the revaluation condition, adapting and choosing optimally after reward revaluation but not after transition revaluation. Specifically, during the learning phase, an SR learner would learn the successor states for each initial state (the SR itself, i.e. $s_1 \rightarrow s_5$; $s_2 \rightarrow s_6$). In the reward revaluation condition, it would then update its reward estimates ($\hat{R}$) for the terminal states ($s_5$ \$1; $s_6$ \$10) during the revaluation phase, much like the model-based learner. Then, during the test phase, it would combine the updated reward estimates with the SR to compute the updated values of the initial states ($s_1 \rightarrow s_5$ \$1; $s_2 \rightarrow s_6$ \$10), allowing it to choose the better one ($s_2$). In contrast, in the transition revaluation condition, the SR learner would not have an opportunity to update the SR of the initial states ($s_1$ and $s_2$) since they are never presented during the revaluation phase, much like the model-free learner. Then, during the test phase, it would combine the unchanged reward estimates with its old but now incorrect SR to produce incorrect estimates for the initial states ($s_1 \rightarrow s_5$ \$10; $s_2 \rightarrow s_6$ \$1) and choose the worse one ($s_1$). 

The pattern of human responses showed evidence of both model-based and SR learning: participants were sensitive to both reward and transition revaluations, consistent with model-based learning, but they performed significantly better after reward revaluations, consistent with SR learning. To rule out the possibility that this effect can be attributed to pure model-based learning with different learning rates for reward ($\hat{R}$) versus transition ($\hat{T}$) estimates, the researchers extended this Pavlovian design to an instrumental design in which participants’ choices (i.e., their policy $\pi$) altered the trajectories they experienced. Importantly, this would correspondingly alter the learned SR: unrewarding states would be less likely under a good policy and hence not be prominent (or not appear at all) in the SR for that policy. Such states could thus get overlooked by an SR learner if they suddenly became rewarding. This subtle kind of reward revaluation (dubbed \textit{policy revaluation} by the researchers) also relies on changes in the reward structure $R$, but induces predictions similar to the transition revaluation condition: SR learners would not adapt quickly, while model-based learners would adapt just as quickly as in the regular reward revaluation condition. 

Human responses on the test phase after policy revaluation were similar to responses after transition revaluation but significantly worse than responses after reward revaluation, thus ruling out a model-based strategy with different learning rates for $\hat{R}$ and $\hat{T}$. Overall, the authors interpreted their results as evidence of a hybrid model-based-SR strategy, suggesting that the human brain can adapt to changes in the environment both by updating its internal model of the world and by learning and leveraging its cached predictive representations \citep[see also][for additional human behavioral data leading to similar conclusions]{kahn2023humans}.

\subsection{Multi-task learning} \label{sec:MTRL}

In the previous section, we saw that humans can adapt quickly to changes in the reward structure of the environment (reward revaluation), as predicted by the SR \citep{momennejad2017successor}. However, that theoretical account alone does not fully explain how the brain can take advantage of the SR to make adaptive choices. Here we take this idea further and propose that humans learn successor \textit{features} (SFs; \S\ref{sec:sf}) for different tasks and use something like the GPI algorithm (\S\ref{sec:sf-gpi}) to generalize across tasks with different reward functions \citep{barreto2017successor,barreto2018transfer,barreto2020fast}.

In \citet{tomov2021multi}, participants were presented with different two-step tasks that shared the same transition structure but had different reward functions determined by the reward weights $\mathbf{w}$ (Figure \ref{fig:sr-cogsci}B).  Each state $s$ was associated with different set of features $\cumulant(s)$, which were valued differently depending on the reward weights $\mathbf{w}$ for a particular task. On each training trial,  participants were first shown the weight vector  $\mathbf{w}_\textit{train}$ for the current trial, and then asked to  navigate the environment in order to maximize reward. At the end of the experiment, participants were presented with a single test trial  on a novel task $\mathbf{w}_\textit{new}$.

The main dependent measure was participant behavior on the test task, which was designed (along with the  training tasks, state features, and transitions) to distinguish between several possible generalization strategies. Across several experiments, \citet{tomov2021multi}   found that  participant behavior was consistent with SF and GPI. In particular, on the first (and only) test trial, participants tended to prefer the training policy that performed better on the new task, even when this was not the optimal policy. This ``policy reuse'' is a key behavioral signature of GPI. This effect could not be explained by model-based or model-free accounts. Their results suggest that humans rely on predictive representations from previously encountered tasks to choose promising actions on novel tasks.

\subsection{Associative learning} \label{sec:associative}

RL provides a normative account of associative learning, explaining how and why agents ought to acquire long-term reward predictions based on their experience. It also provides a descriptive account of a myriad of phenomena in the associative learning literature \citep{sutton1990time,niv2009reinforcement,ludvig2012evaluating}. Two recent ideas have added nuance to this story:

\begin{itemize}
    \item \textbf{Bayesian learning}. Animals represent and utilize uncertainty in their estimates.
    \item \textbf{Context-dependent learning}. Animals partition the environment into separate contexts and maintain separate estimates for each context.
\end{itemize}

We examine each idea in turn and then explore how they can be combined with the SR (\S\ref{sec:sr}). In brief, the key idea is that animals learn a probability distribution over context-dependent predictive representations.

\subsubsection{Bayesian RL}
\label{sec:bayesianrl}

While standard RL algorithms (\S\ref{sec:algorithmic}) learn point estimates of different unknown quantities like the transition function ($\hat{T}$) or the value function ($\hat{V}$), Bayesian RL posits that agents treat such unknown quantities as random variables and represent beliefs about them as probability distributions. 

Generally, Bayesian learners assume a generative process of the environment according to which hidden variables give rise to observable data. The hidden variables can be inferred by inverting the generative process using Bayes' rule. In one example of Bayesian value learning \citep{gershman2015unifying}, the agent assumes that the expected reward $R(s,\mathbf{w})$ is a linear combination of the observable state features $\cumulant(s)$:\footnote{Note that the linear reward model is the same as assumed in much of the SF work reviewed above (see Eq. \ref{eq:sf-reward-relationship}).}
\begin{align}
    R(s,\mathbf{w})= \cumulant\left(s\right)^{\top} \task
    \label{eq:linear_reward}
\end{align}
where the hidden weights $\mathbf{w}$ are assumed to evolve over time according to the following dynamical equations:
\begin{align}
    \mathbf{w}_0 &\sim \mathcal{N}(\mathbf{0}, \mathbf{\Sigma}_0) ,\\
    \mathbf{w}_t &\sim \mathcal{N}(\mathbf{w}_{t-1}, q\mathbf{I})  \\
    r_t &\sim \mathcal{N}(R(s_t,\mathbf{w}_t), \sigma^2_R).
\end{align}
In effect, this means that each feature (or stimulus dimension) is assigned a certain reward weight. The weights are initialized randomly around zero (with prior covariance $\mathbf{\Sigma}_0$) and evolve according to a random walk (with volatility governed by the transition noise variance $q$). Observed rewards are given by the linear model (Eq. \ref{eq:linear_reward}) plus zero-mean Gaussian noise with variance $\sigma^2_R$.

This formulation corresponds to a linear-Gaussian dynamical system (LDS). The posterior distribution over weights given the observation history $H_t = (s_{1:t},r_{1:t})$ follows from Bayes' rule:
\begin{align}
    p(\mathbf{w}_t|H_t) = \frac{p(H_t|\mathbf{w}_t) p(\mathbf{w}_t)}{p(H_t)} = \mathcal{N}(\mathbf{w}_t; \hat{\mathbf{w}}_t, \mathbf{\Sigma}_t).
\end{align}
The posterior mean $\hat{\mathbf{w}}_t$ and covariance $\mathbf{\Sigma}_t$ can be computed recursively in closed form using the Kalman filtering equations:
\begin{align}
    \hat{\mathbf{w}}_t &= \hat{\mathbf{w}}_{t-1} + \mathbf{k}_t \delta_t   \label{eq:kalmanmean} \\   
    \mathbf{\Sigma}_t &= \mathbf{\Sigma}_{t-1} + q\mathbf{I} - \lambda_t \mathbf{k} \mathbf{k}_t^\top   \label{eq:kalmanvar}  ,
\end{align}
where
\begin{itemize}
    \item $\delta_t = r_t - \cumulant(s_t)^\top  \hat{\mathbf{w}}_t$ is the reward prediction error.
    \item $\lambda_t = \cumulant(s_t)^\top (\mathbf{\Sigma}_t + q\mathbf{I}) \cumulant(s_t) + \sigma_R^2$ is the residual variance.
    \item $\mathbf{k}_t = (\mathbf{\Sigma}_t + q\mathbf{I}) \cumulant(s) / \lambda$ is the Kalman gain.
\end{itemize}
This learning algorithm is generalizes the seminal Rescorla-Wagner model of associative learning \citep{rescorla1972theory}, and its update rule bears resemblance to the error-driven TD update (\Eqref{eq:Qlearn}). However, there are a few notable distinctions from its non-Bayesian counterparts:
\begin{itemize}
    \item \textbf{Uncertainty-modulated updating}. The learning rate corresponds to the Kalman gain $\mathbf{k}_t$, which increases with posterior uncertainty (the diagonals of the posterior covariance, $\mathbf{\Sigma}_t$).
    \item \textbf{Non-local updating}. The update is multivariate, affecting the weights of all features simultaneously according to $\mathbf{k}_t$. This means that weights of ``absent'' features (i.e., where $\phi_i(s) = 0$) can be updated, provided those features have non-zero covariance with observed features.
\end{itemize}
These properties allow the Kalman filter to explain a number of phenomena in associative learning that elude non-Bayesian accounts \citep{dayan2000explaining,kruschke2008bayesian,gershman2015unifying}. Uncertainty-modulated updating implies that nonreinforced exposure to stimuli, or even just the passage of time, can impact future learning. For example, in a phenomenon known as \emph{latent inhibition}, pre-exposure to a stimulus reduces uncertainty, which in turn reduces the Kalman gain, retarding subsequent learning for that stimulus. Non-local updating implies that learning could occur even for unobserved stimuli, if they covary with observed stimuli according to $\mathbf{\Sigma}_t$. For example, in a phenomenon known as \emph{backward blocking}, reinforcing two stimuli simultaneously (a compound stimulus, $\cumulant(s) = [1, 1]$) and subsequently reinforcing only one of the stimuli ($\cumulant(s') = [1, 0]$) reduces reward expectation for the absent stimulus ($\cumulant(s'') = [0, 1]$) due to the learned negative covariance between the two reward weights ($\mathbf{\Sigma}_{1,2}$ < 0). 

Kalman filtering can also been extended to accommodate long-term reward expectations in an algorithm known as Kalman TD \citep{geist2010kalman} by replacing the state features $\cumulant(s)$ with their discounted temporal difference:
\begin{align}
    \mathbf{h}_t = \cumulant(s_t) - \gamma \cumulant(s_{t+1}) . \label{eq:kalmanTD}
\end{align}
This recovers the Kalman filter model described above when the discount factor $\gamma$ is set to 0. Importantly, this model provides a link between the Bayesian approach to associative learning and model-free reinforcement learning algorithms; we exploit this link below when we discuss Bayesian updating of beliefs about predictive representations. \citet{gershman2015unifying} showed that Kalman TD can explain a range of associative learning phenomena that are difficult to explain by Bayesian myopic ($\gamma=0$) models and non-myopic ($\gamma>0$) non-Bayesian models.

\subsubsection{Context-dependent learning}
\label{sec:context}

In addition to accounting for uncertainty in their estimates, animals have been shown to learn different estimates in different contexts \citep{redish2007reconciling}. This kind of context-dependent learning can also be cast in a Bayesian framework \citep{courville2006bayesian,gershman2010context} by assuming that the environment is carved into separate contexts according to a hypothetical generative process. One possible formalization is the \emph{Chinese restaurant process} (CRP), in which each observation is conditioned on a hidden context $z_t$ that is currently active and evolves according to:
\begin{align}
    P(z_t = k \mid \mathbf{z}_{1:t-1}) \propto \begin{cases}
    N_k \,\,\,\,\,    &\text{if $k$ is a previously sampled context} \\
 \alpha \,\,\,\,\, &\text{otherwise}
\end{cases}  \label{eq:CRP}
\end{align}
where $N_k = \sum_{i=1}^{t-1} \mathbb{I}[z_i = k]$ is the number of previous time steps assigned to context $k$ and $\alpha \geq 0$ is a \emph{concentration parameter}, which can be understood as controlling the effective number of contexts.\footnote{Under the CRP, the expected number of contexts after $t$ time steps is $\alpha \log t$.} 

If each context is assigned its own probability distribution over observations, then inferring a given context $k$ is driven by two factors:
\begin{itemize}
    \item How likely is the current observation under context $k$?
    \item How likely is context $k$ given the previous context assignments $\mathbf{z}_{1:t-1}$ (\Eqref{eq:CRP})?
\end{itemize}
In particular, a given context is more likely to be inferred if the current observations are more likely under its observation distribution and/or if it has been inferred more frequently in the past (that is, if past observations have also been consistent with its observation distribution). Conversely, if the current observations are unlikely under any previously encountered context, a new context is induced with its own observation distribution. 

This formulation has accounted for a number of phenomena in the animal learning literature \citep{gershman2010context}. For example, it explains why associations that have been extinguished sometimes reappear when the animal is returned to the context in which the association was first learned, a phenomenon known as \emph{renewal}. It also provides an explanation of why the latent inhibition effect is attenuated if pre-exposure to a stimulus occurs in one context and reinforcement occurs in a different context.  

Recently, a similar formulation has been used to explain variability in the way hippocampal place cells change their firing patterns in response to changes in contextual cues, a phenomenon known as \emph{remapping} \citep{sanders2020hippocampal}. On this view, the hippocampus maintains a separate cognitive map of the environment for each context, and hippocampal remapping reflects inferences about the current context.

\subsubsection{Bayesian learning of context-dependent predictive maps}

As discussed in \S\ref{sec:hippocampus}, \citet{stachenfeld2017hippocampus} showed that many aspects of hippocampal place cell firing patterns are consistent with the SR, suggesting that the hippocampus encodes a predictive map of the environment. In this light, the view that the hippocampus learns different maps for different contexts \citep{sanders2020hippocampal} naturally points to the idea of a context-dependent predictive map.  

This idea was formalized by \citet{geerts2023probabilistic} in a model that combines SFs (\S\ref{sec:sf}) with Bayesian learning and context-dependent learning. Their model learns a separate predictive map of the environment for each context, where each map takes the form of a probability distribution over SFs (Figure~\ref{fig:sr-cogsci}C). The generative model assumes that the SF is given by a linear combination of state features:
\begin{align}
    \sf_{j,t}(s) = \cumulant(s)^\top \mathbf{m}_{j,t},
\end{align}
where $\mathbf{m}_j$ is a vector of weights for predicting successor feature $j$. Analogous to the Kalman TD model described above, Geerts et al. assume that the weight vectors for all features evolve over time according to the following LDS:
\begin{align}
    \mathbf{m}_{j,0} &\sim \mathcal{N}(\mathbf{\mu}_0, \mathbf{\Sigma}_0)  \\
    \mathbf{m}_{j,t} &\sim \mathcal{N}(\mathbf{m}_{j,t-1}, q\mathbf{I})  \\
    \phi_{j,t}(s_t) &\sim \mathcal{N}(\mathbf{h}_t^\top \mathbf{m}_{j,t-1} , \sigma_\phi^2). \label{eq:kalmanSRobs}
\end{align}
This means that the weight vector is initialized randomly (with prior mean $\mathbf{\mu}_0$ and variance $\mathbf{\Sigma}_0$) and evolves randomly according to a random walk (with transition noise variance $q$). Observed features $\cumulant(s)$ are noisy differences in the SFs of successive states (with observation noise variance $\sigma_\phi^2$). As in Kalman TD, the posterior over the weight vector for feature $j$ is also Gaussian with mean $\hat{\mathbf{m}}_j$ and variance $\mathbf{\Sigma}$ which can be computed using Kalman filtering equations essentially the same as those given above (\Eqref{eq:kalmanmean}-\ref{eq:kalmanvar} in \S\ref{sec:bayesianrl}).

The authors generalize this to multiple contexts using a \emph{switching LDS}: each context has a corresponding LDS and SF weights. The contexts themselves change over time based on a ``sticky'' version of the Chinese restaurant process, which additionally favors remaining in the most recent context:
\begin{align}
    P(z_t = k \mid \mathbf{z}_{1:t-1}) \propto \begin{cases}
                                            N_k + \nu \mathbb{I}[z_{t-1} = k] \,\,\,\,\,    &\text{if $k$ is a previously sampled context} \\
                                            \alpha  \,\,\,\,\, &\text{otherwise}
                                         \end{cases},  \label{eq:sCRP}
\end{align}
where $\nu \geq 0$ is the ``stickiness'' parameter dictating how likely it is to remain in the same context. 

Thus a given context $k$ is more likely to be inferred if:
\begin{itemize}
    \item The current observations are consistent with its SFs (\Eqref{eq:kalmanSRobs}).
    \item It was encountered frequently in the past (driven by the $N_k$ term in~\Eqref{eq:sCRP}).
    \item It was encountered recently (driven by the $\mathbb{I}[z_{t-1} = k]$ term in~\Eqref{eq:sCRP}).
\end{itemize}
Conversely, a new context is more likely to be inferred if observations are inconsistent with the current SFs (i.e., when there are large SF prediction errors).

The authors show that this model can account for a number of puzzling effects in the animal learning literature that pose problems for both point estimation (TD learning) of the SR/SF (\Eqref{eq:SRTDupdate}-\ref{eq:SRTDerror} in \S\ref{sec:sr}) and Bayesian RL (\Eqref{eq:kalmanmean}-\ref{eq:kalmanvar} in \S\ref{sec:bayesianrl}). One example is the opposing effect that pre-exposure to a context can have on learning. Brief exposure to a context can facilitate learning (context pre-exposure facilitation), while prolonged exposure can inhibit learning \citep[latent inhibition;][]{kiernan1993effects}. Context pre-exposure facilitation by itself can be accounted for by TD learning of the SR alone \citep{stachenfeld2017hippocampus}: during pre-exposure, the animal learns a predictive representation which facilitates propagation of newly learned values. Latent inhibition by itself can be accounted for by Bayesian RL alone \citep{gershman2015unifying}, as discussed previously: prolonged exposure reduces value uncertainty, in turn reducing the Kalman gain $\mathbf{k}_t$ (the effective learning rate) and inhibiting learning of new values. Kalman learning of SFs combines these two processes and can thus resolve the apparent paradox: initially, the animal learns a predictive representation of the context, which facilitates learning, whereas after prolonged exposure, this effect is offset by a reduction in value uncertainty, which inhibits learning.

Another puzzling effect is the partial transition revaluation observed in \citet{momennejad2017successor} and discussed in \S\ref{sec:revaluation}, which cannot be accounted for by TD learning of the SR. This led Momennejad et al. to propose a hybrid model-based-SR strategy that relies on offline simulations. Kalman TD offers a more parsimonious account based on non-local updating that does not appeal to model-based simulations. In particular, the covariance matrix $\mathbf{\Sigma}_t$ learned during the learning phase captures the relationship between the initial states ($s_1$ and $s_2$) and subsequent states ($s_3$ and $s_4$). Updating the transitions from those subsequent states ($s_3 \rightarrow s_6$ and $s_4 \rightarrow s_5$) during the transition revaluation phase therefore also updates the SR for the initial states ($s_1$ and $s_2$), even though they are not encountered during the revaluation phase.

Non-local updating can similarly explain reward devaluation of preconditioned stimuli, a hallmark of model-based learning \citep{hart2020responding}. This is similar to reward devaluation discussed in \S\ref{sec:revaluation}, except with an additional preconditioning phase during which an association is learned between two neutral stimuli (e.g., light $\rightarrow$ tone). During the subsequent conditioning phase, the second stimulus is associated with a rewarding outcome (e.g., tone $\rightarrow$ food), which is then devalued (e.g., by inducing taste aversion) during the devaluation phase. Finally, the animal is tested on the first neutral stimulus (e.g., light). Note that since the first stimulus is never present during the conditioning phase, TD learning would not acquire an association between the first stimulus and the reward, and would thus not exhibit sensitivity to reward devaluation \citep{gardner2018rethinking}. In contrast, during the preconditioning phase, Kalman TD learns that the two stimuli covary, allowing it to update the SF for both stimuli during the conditioning phase and consequently propagate the updated value to both stimuli during the devaluation phase.

Note that the phenomena so far can be explained without appealing to context-dependent learning (\S\ref{sec:context}), since the experiments take place in the same context. Context-dependent Kalman TD can additionally explain a number of intriguing phenomena when multiple contexts are introduced. 

One such phenomenon is the context specificity of learned associations \citep{winocur2009changes}. In this paradigm, an animal learns an association (e.g., tone $\rightarrow$ shock) in one context (e.g., Context A) and is then tested in the same context or in another context (e.g., Context B). The amount of generalization of the association across contexts was found to depend on elapsed time: if testing occurs soon after training, the animal only responds in the training context (A), indicating context specificity. However, if testing occurs after a delay, the animal responds equally in both contexts A and B, indicating contextual generalization. Even more intriguingly, this effect is reversed if the animal is briefly reintroduced to the training context (A) before testing, in which case responding is once again context-specific---a hippocampus-dependent ``reminder'' effect.

The context-dependent model readily accounts for these effects. Shortly after training, the uncertainty of the SR assigned to Context A is low (i.e., the animal is confident in its predictive representation of Context A). Introduction to Context B therefore results in a large prediction error, leading the animal to (correctly) infer a new context with a new SR, leading to context-specific responding. However, as more time elapses, the uncertainty of the SR assigned to Context A gradually increases (i.e., the animal becomes less confident in its predictive representation of Context A, a kind of forgetting). Introduction to Context B then results in a smaller prediction error, making it likely that the new observations are also assigned to Context A, leading to generalization across contexts. Brief exposure to Context A reverses this effect by reducing the uncertainty of the SR assigned to Context A (i.e., the animal's confidence in its predictive representation of Context A is restored, a kind of remembering), leading once again to context-specific responding.

Recall that the duration of context pre-exposure has opposing effects on learning, initially facilitating but subsequently inhibiting learning \citep{kiernan1993effects}. But what if the animal is tested in a different context? In a follow-up experiment, \citet{kiernan1993effects} showed that longer pre-exposure to the training context leads to less responding in the test context, indicating that the learned association is not generalized. That is, longer context pre-exposure has a monotonic inhibitory effect on generalization across contexts. The context-dependent model can account for this with the same mechanism that accounts for context pre-exposure facilitation: longer pre-exposure to the training context reduces the uncertainty of its predictive estimate, leading to greater prediction errors when presented with the text context, and increasing the probability that the animal will infer a new context, leading to context-specific responding.

Overall, the results of \citet{geerts2023probabilistic} suggest that, rather than encoding a single monolithic predictive map of the environment, the hippocampus encodes multiple separate predictive maps, each associated with its own context. Both the context and the predictive map are inferred using Bayesian inference: learning of the predictive map is modulated by uncertainty, and supports non-local updating. A new context is inferred when the current predictive map fails to account for current observations.

\subsection{Spatial navigation}  \label{sec:spatial}

A rich body of work points to the hippocampus as encoding a kind of cognitive map of the environment that mammals rely on for navigation in physical and abstract state spaces \citep{o1971hippocampus,okeefeHippocampusCognitiveMap1978}. As we discussed in the previous section and in \S\ref{sec:hippocampus}, this cognitive map can be usefully interpreted as a predictive map in which states predict future expected states, consistent with the SR \citep[\S\ref{sec:sr};][]{stachenfeld2017hippocampus}. Yet despite many studies of navigation in humans and rodents---key model species used to study spatial navigation \citep{epstein2017cognitive,ekstrom2018space,ekstrom2018human,gahnstrom2020striatal,nyberg2022spatial,spiers2015neural}---until recently there was no direct comparison of human and rodent navigation in a homologous task. This left open the question of whether spatial navigation across mammalian species relies on an evolutionarily conserved strategy supported by such a predictive map. 

A recent study by \citet{de2022predictive}  filled this gap  by designing a homologous navigation task for humans and rats. They devised a configurable open-field maze that could be reconfigured  between trials, allowing experimenters to assess a hallmark aspect of spatial navigation: the ability to efficiently find detours and shortcuts. The maze consisted of a 10-by-10  grid in which squares could be blocked off by the experimenters. The maze was instantiated  in a physical environment for rats and in a virtual reality environment for humans. 

On each trial, the participant was placed at a starting location  and had to navigate to a goal location to receive a reward (Figure~\ref{fig:sr-cogsci}D).  The starting location varied across trials, while the goal remained hidden at a fixed location throughout the experiment.  Keeping the goal location unobservable ensured that  participants could not rely on simple visual heuristics (e.g.,  proximity to the goal).  At the same time,  keeping the goal location fixed ensured that, once  it is identified,  the key problem becomes navigating to it,  rather than re-discovering it. During the training phase  of the experiment,  all squares of the grid were accessible,  allowing participants to learn an internal map of the environment.  During the test phase,  participants were  sequentially presented with 25  different maze configurations  in which various sections of the maze were blocked off. Participants completed 10 trials of each configuration before moving on to the next. 

Using this task, the authors compared human and  rat navigation  with three types of RL  algorithms:
\begin{itemize}

\item \textbf{Model-free  agent} (\S\ref{sec:algorithmic}).  No internal map of the environment; optimal policy is based on state-action value function $Q$, which is learned from experience (specifically, using Q-learning (\Eqref{eq:Qlearn}) with eligibility traces).

\item \textbf{Model-based agent} (\S\ref{sec:algorithmic}).  Full internal map of the environment (transition structure $T$  and reward function $R$) is learned from experience;  optimal policy is computed using tree search at  decision time (specifically, using A* search).

\item 	\textbf{SR agent} (\S\ref{sec:sr}). Predictive map of the environment (SR matrix $\mathbf{M}$ and  reward function $R$) is learned from experience; optimal policy is computed by combining SR and  reward function (\Eqref{eq:sr-value}).

\end{itemize}
The key question that the authors sought to answer was which RL  strategy best  explains human and rat  navigation across the novel test configurations. Across a wide range of analyses, the authors observed a consistent trend: both human and rodent behavior was most consistent with the SR agent. Humans also showed  some  similarity to the model-based agent, but neither species was consistent with the model-free agent.

First, the authors simulated each RL agent generatively on the same trials as the participants---i.e., they let the RL agent navigate and solve each trial as a kind of simulated participant, learning from its own experience along the way. These closed-loop\footnote{We refer to the them as ``closed-loop'' since, at each step, the  output of the RL agent (its action) is fed back to change its position on the grid, influencing the agent’s input at the following step, and so on.} simulations show what overall participant behavior would look like according to each RL strategy. This revealed that:
\begin{itemize}
\item	Model-free agents struggle  on new maze configurations due to the slow learning of the $Q$-function, which takes many trials to propagate values from the goal location to possible starting locations.
\item	Model-based agents generalize quickly to new maze configurations, since local updates to the transition structure $T$ can be immediately reflected in the tree search algorithm.
\item	SR agents generalize faster than model-free but more slowly than model-based agents, since updates to the SR matrix $\mathbf{M}$ reach farther than updates to the $Q$-function, but still require several trials to propagate all the way to the possible starting locations.
\end{itemize}

Second, the authors clamped each RL agent to participant behavior---i.e., they fed the agent the same sequence of states and actions experienced by a given participant.  These open-loop simulations show what the participant would do  at each step if they were following a given RL strategy.\footnote{We refer to them as ``open-loop''  since, at each step, the output of the RL agent has no effect on its subsequent inputs or outputs.} By matching these predictions with participant behavior using maximum likelihood estimation of model parameters, the authors  quantified how consistent step-by-step participant behavior is with each RL strategy. For both humans and rats,  this analysis revealed the greatest similarity (i.e., highest likelihood) with the SR agent, followed by the model-based agent, with the model-free agent showing the least similarity (i.e., lowest likelihood).

Third, the authors combined the above approaches by first training each fitted RL agent with the state-action sequences observed by a participant on several maze configurations and then simulating it generatively on another configuration. This hybrid open-loop training (on past configurations) / closed-loop evaluation (on a new configuration) provides a more global view than the step-by-step analysis above, by allowing comparison of predicted and participant \textit{trajectories}, rather than  individual \textit{actions}. This led to several findings:
\begin{itemize}
\item Configurations that were challenging for the SR  agent were also challenging for biological agents, and vice versa. This pattern was less consistent for model-based and model-free agents.
\item Overall directedness and direction of participant trajectories (quantified by linear and angular diffusivity) was most similar to SR trajectories.
\item The step-by-step distance between participant and SR  trajectories was consistently lower compared to model-based and model-free trajectories.
\end{itemize}

All of these analyses show that the SR agent best explains both human and rat behavior. Overall, the results of \citet{de2022predictive} indicate that spatial navigation across mammalian species relies on a predictive map that is updated from experience in response to changes in the environment.

\subsection{Memory}  \label{sec:memory}

The hippocampus and the adjacent medial temporal lobe structures are also involved in another high-level cognitive function: episodic memory \citep{ranganath10}. In this section, we review an influential model of episodic memory, the \emph{Temporal Context Model} \citep[TCM;][]{howard2002distributed}, through the lens of RL, and show that it can be partially understood as an estimator for the SR \citep[\S\ref{sec:sr};][]{gershman2012successor}. We then discuss how this property can be used in a powerful decision-making algorithm that bridges episodic memory and reinforcement learning systems \citep{zhou2023episodic}. 

\subsubsection{The Temporal Context Model}
\label{sec:tcm}

TCM is an influential model of memory encoding and retrieval originally designed to account for a number of phenomena in free recall experiments \citep{howard2002distributed}. In these experiments, participants are asked to study a list of items and then recall as many of them as they can, in any order. Experimenters observed that recall order is often not, in fact, arbitrary:  participants  show better recall for recently studied items (the recency effect), and tend to recall adjacent items in the list one after the other (the contiguity effect). %

TCM accounts for these phenomena by positing that the brain maintains a \textit{temporal context}: a slowly drifting internal representation of recent experience that gets bound to specific experiences (memories) during encoding, and can serve as a cue to bring those experiences to mind during retrieval. When participants begin recalling the studied items, the temporal context is most similar to the context associated with recently studied items (due to the slow drift), which is why they are recalled better (the recency effect). Recalling items reactivates the context associated with those items, which is similar to the context for adjacent items (again, due to the slow drift), which is why they tend to be recalled soon after (the contiguity effect). Neural evidence for drifting context comes from the finding that lingering brain activity patterns of recent stimuli predicted whether past and present stimuli are later recalled together \citep{chan2017lingering}. Human brain recordings have also provided evidence for temporal context reactivation during recall \citep{gershman2013neural,folkerts2018human}.

Temporal context can be formalized as weighted average of recently encountered stimulus vectors:
\begin{align}
    \mathbf{c}_{t+1} = (1 - \omega) \mathbf{c}_t + \omega \cumulant(s_t), \label{eq:TCM}
\end{align}
where $\mathbf{c}_t$ is the  current context vector, $\cumulant(s_t)$ is the feature vector for the current stimulus $s_t$, which could correspond to a study item (in an episodic memory setting) or a state (in a decision making setting), and the constant $\omega$   determines the drift rate---i.e., whether the context evolves slowly (low $\omega$) or quickly (high $\omega$). 

Stimuli and contexts are bound using an association matrix $\hat{\mathbf{M}}$  which gets updated using outer-product Hebbian learning: when a new stimulus is presented, its associations with previous stimuli are strengthened in proportion to how active they are in the current context:
\begin{align}
   \Delta \hat{M}(i,j) \propto \phi_i(s_t) c_{i,t}.   \label{eq:TCMupdate}
\end{align}
In early studies of this model, stimuli were encoded using one-hot vectors, $\phi_i(s) = \mathbb{I}[s=i]$, although more flexible feature representations have been studied \citep[e.g.,][]{socher2009bayesian,howard2011constructing}.

\subsubsection{TCM as an estimator for the SR}

Note the similarity between the TCM learning rule (\Eqref{eq:TCMupdate}) and the SR TD learning rule (\Eqref{eq:SRTDupdate} in \S\ref{sec:sr}). There are two main  distinctions:
\begin{itemize}
\item	 The TCM update (\Eqref{eq:TCMupdate}) is modulated by the context vector $\mathbf{c}$,  which is absent from the SR TD update (\Eqref{eq:SRTDupdate} in \S\ref{sec:sr}).
\item	The Hebbian update in TCM lacks the prediction error terms from the SR TD error $\delta_M(j)$ (\Eqref{eq:SRTDerror} in \S\ref{sec:sr}).
\end{itemize}
The first distinction can be removed by setting a maximum drift rate of $\omega = 1$ in~\Eqref{eq:TCM}, which ensures that the context is always updated to the latest stimulus vector, $\mathbf{c}_{t+1} = \cumulant(s_t)$. For the one-hot encoding, this means that only $\hat{M}(s_{t-1},j)$ will be updated in the TCM update, since $c_{i,t} = \mathbb{I}[s_{t-1}=i]$ in~\Eqref{eq:TCMupdate}), just as in the SR TD update (\Eqref{eq:SRTDupdate}). Conversely, introducing a context term in the SR TD  update (\Eqref{eq:SRTDupdate}) results in a generalization of TD learning using an \emph{eligibility trace} \citep{sutton18}, a running average of recently visited states. This is mathematically equivalent to temporal context (\Eqref{eq:TCM}), and can sometimes lead to faster convergence. In this way, the TCM learning rule is a generalization of the vanilla SR TD learning rule that additionally incorporates eligibility traces.

The second distinction suggests a way to, in turn,  generalize the TCM update rule by replacing the Hebbian term with the prediction error term $\delta_M(j)$ (\Eqref{eq:SRTDerror}) from the SR TD update. This leads to the following revised update equation for TCM:
\begin{align}
    \Delta \hat{M}(i,j) \propto \delta_M (j) c_{i,t},   \label{eq:TCMTDupdate}
\end{align}
where $\delta_M(j)$ is the same as in~\Eqref{eq:SRTDerror}. \citet{gershman2012successor} showed that this new variant of TCM can be understood as directly estimating the SR using TD learning, with temporal context serving as an eligibility trace. It differs from the original version of TCM in two keys ways. First,  learning is error-driven rather than purely Hebbian, which means that association strength only grows if there is a discrepancy between predicted and observed stimuli \citep[see][for a discussion of empirical evidence]{dubrow2017does}. Second, associations are additionally learned between the context and future expected stimuli, not just the present stimulus.

This new interpretation of TCM posits that the role of temporal context is to learn predictions of future  stimuli, rather than to merely form associations. This makes several distinct predictions from the original version of TCM, one of which is the context repetition effect: repeating the context in which a stimulus was observed should strengthen memory for that stimulus, even if the stimulus itself was not repeated. This prediction was validated in a study by \citet{smith2013context}. The authors showed participants triplets of stimuli (images in one experiment and words in another experiment), with the first two stimuli in each triplet serving as context for the third stimulus.  Participants were then presented with an item-recognition test in which they had to indicate whether different stimuli are either ``old'' or ``new''. Memory performance was quantified as the proportion of test items correctly recognized as old. The key finding was that memory was better for stimuli whose context was presented repeatedly,  even if the stimuli themselves were  only presented once.   This held for different modalities (images and words), and did not occur when context was generally not predictive of stimuli. These findings \citep[see also][]{manns2015temporal} substantiate the predicted context repetition effect and lend credence to idea that TCM learns predictions rather than mere associations.

\subsubsection{Combining TCM and the SR for decision making}

The theoretical links between TCM (\S\ref{sec:tcm}) and the SR (\S\ref{sec:sr}) point to a broader role for episodic memory in RL. Previous studies have implicated the hippocampus in prediction and imagination \citep{buckner2010role}, as well as replay of salient events \citep{momennejad2018offline}, consistent with some form of model-based RL or a successor model (SM; \S\ref{sec:successor-models}). More generally, episodic memory is thought to support decision making by providing the ingredients for simulating possible futures \citep{schacter2015episodic}. This idea is corroborated by studies of patients with episodic memory deficits, who also tend to show deficits on decision making tasks \citep{gupta2009declarative,gutbrod2006decision,bakkour2019hippocampus}. A related body of work focuses on decision-by-sampling algorithms, according to which humans approximate action values by sampling from similar past experiences stored in memory \citep{stewart2006decision,plonsky2015reliance,bornstein2017reminders,lieder2018overrepresentation,bhui2018decision}.

These loosely connected ideas were knitted together in a theoretical proposal by \citet{zhou2023episodic}  that builds upon the links between TCM and the SR, showing precisely how a predictive version of TCM can support adaptive decision making. Their model incorporates two key ideas (Figure~\ref{fig:sr-cogsci}E):
\begin{itemize}
\item	 During encoding (Figure~\ref{fig:sr-cogsci}E, top), incoming feature vectors $\cumulant(s_t)$ update a slowly drifting context $\mathbf{c}_t$ (\Eqref{eq:TCM}). This context vector serves as an eligibility trace in a TD update rule (\Eqref{eq:TCMTDupdate}) that learns the SR estimate $\hat{\mathbf{M}}$ \citep{gershman2012successor}. 
\item	  During retrieval (i.e., at decision time; Figure~\ref{fig:sr-cogsci}E, middle and bottom), possible future stimuli  $\mathbf{x}$  are sampled for each action using a tree-search algorithm that uses the SR as a world \textit{model}, effectively turning it into a SM (\S\ref{sec:successor-models}): $p(\tilde{s}_\tau) \propto  \hat{\mathbf{M}} \mathbf{c}_\tau$, where $\tau$ indexes time steps at retrieval, and $\tilde{s}_0$ corresponds to the initial state of the retrieval process (the query stimulus, defined as the root of the tree). Retrieval unfolds by recursively sampling states from this process. The corresponding rewards then are averaged to compute a Monte Carlo value estimate for each action.
\end{itemize}
During the tree search, the context vector $\mathbf{c}_\tau$ can be updated with the sampled feature vector $\cumulant(s_t)$  to varying degrees, dictated by the drift rate $\omega$ (\Eqref{eq:TCM}). This spans a continuum between updating and retrieval regimes.

At one extreme, if the drift rate during retrieval is set to its lowest value ($\omega = 0$), the context is never updated after being initialized with the query stimulus ($\mathbf{c}_\tau = \mathbf{x}_0 = \cumulant(\tilde{s}_0) \forall \tau$). This results in independent and identically distributed samples from the normalized SR (i.e. the SM). In the limit of infinite samples,  this reduces to simply computing action values by combining the reward function and the SR (\Eqref{eq:sr-value} in \S\ref{sec:sr}),  as  discussed in \S\ref{sec:revaluation} and \S\ref{sec:MTRL}. For finite samples, this produces an unbiased estimator of action values. Note that this estimate is only as accurate as the SR matrix $\hat{\mathbf{M}}$, which is itself an estimate of the true SR matrix $\mathbf{M}$. Hence, this regime inherits all the pros and cons of using the SR (\S\ref{sec:sr} and \S\ref{sec:revaluation}): it can be used to  efficiently compute action values, and it can adapt quickly to changes in the reward structure $R$, but not the transition structure $T$ of the environment.

At the other extreme, if the drift rate during retrieval is set to its highest value ($\omega = 1$), the context is always updated to the latest sampled stimulus ($\mathbf{c}_\tau = \cumulant(\tilde{s}_\tau)$). If the discount factor is minimal, $\gamma = 0$, the SR reduces to the one-step transition matrix (i.e., $\mathbf{M} = \mathbf{T}$)  and the sampled stimuli $s_\tau$ are no longer independent and identically distributed, but instead form a trajectory through state space that follows the transition structure $T$ and corresponds to a single Monte Carlo rollout. Averaging rewards from such Monte Carlo rollouts  also produces an unbiased estimator of value \citep{sutton18}. This regime thus corresponds to a fully model-based algorithm (\S\ref{sec:algorithmic}) and inherits all the pros and cons of that approach: it takes longer to compute action values (since trajectories need to be fully rolled out to produce unbiased estimates, requiring more samples), but it can adapt quickly to changes in both the reward structure $R$ and the transition structure $T$ of the environment. 

Between these extremes lies a continuum ($0 < \omega < 1$)  that trades off between a sampling approximation of the SR ($\omega \rightarrow 0$) and model-based Monte Carlo rollouts ($\omega \rightarrow 1$). Indeed, results from free recall experiments are consistent with such an intermediate regime \citep{howard2002distributed}, indicating that context is partially updated during retrieval. This also raises the intriguing possibility that the brain navigates this continuum by dynamically adjusting the drift rate in a way that balances the pros and cons of both regimes,  similarly to the way in which the brain arbitrates between model-based and model-free RL systems \citep{kool2018competition}.

The authors also demonstrate how emotionally salient stimuli, such as high rewards, can modulate learning by  producing a higher learning rate for the SR update (\Eqref{eq:TCMTDupdate}). This introduces a kind of bias-variance trade-off: the resulting SR skews towards stimuli that were previously rewarded, which could speed up convergence (lower variance) but also induce potentially inaccurate action values (higher bias). Finally,  the authors show how initiating the tree search with a retrieval context $\mathbf{c}_0$ that is associated with but different from the query stimulus feature vector $\cumulant(\tilde{s}_0)$  can lead to bidirectional retrieval.  This is consistent with bidirectional recall in human memory experiments, and can be advantageous in problems where state transitions can be bidirectional, such as spatial navigation.

In summary, the modeling and simulation results of \citet{zhou2023episodic} demonstrate how a variant of TCM can be viewed as an estimator of the SR, and can serve as the basis for a flexible sampling-based decision algorithm that spans the continuum between vanilla SR and fully model-based search. This work illustrates how episodic memory can be integrated with predictive representations to explain cognitive aspects of decision making.%

\section{Conclusions}

The goal of this survey was to show how predictive representations can serve as the building blocks for intelligent computation. Modern work in AI has demonstrated the power of good representations for a variety of downstream tasks; our contribution builds on this insight, focusing on what makes representations useful for RL tasks. The idea that representing predictions is particularly useful has appeared in many different forms over the last few decades, but has really blossomed only in the last few years. We now understand much better \emph{why} predictive representations are useful, \emph{what} they can be used for, and \emph{how} to learn them.

\subsection*{Acknowledgments}

This work has been made possible in part by a gift from the Chan Zuckerberg Initiative Foundation to establish the Kempner Institute for the Study of Natural and Artificial Intelligence, an ARO MURI under Grant Number W911NF-23-1-0277, and the Wellcome Trust under Grant Number 212281/Z/18/Z.

\bibliographystyle{apalike}
\bibliography{bib}

\clearpage

\end{document}